\documentclass[journal]{IEEEtran}
\usepackage{pdfpages}
 \pdfoutput=1
\usepackage{graphicx}
\usepackage{bm}
\usepackage{amsmath}
\usepackage{algorithm}
\usepackage{algorithmic}
\usepackage{pifont}
\usepackage{algorithm}
\usepackage{color}
\usepackage{amsfonts}
\usepackage{amsmath}
\usepackage{multirow}
\usepackage{threeparttable} 
\usepackage{chngpage}
\usepackage{epstopdf}
\usepackage{epsfig}
\usepackage{arydshln}
\usepackage{fixltx2e}
\usepackage{bbm}
\usepackage{scrextend}

\newcommand{\argmin}{\operatornamewithlimits{argmin}}

\makeatletter
\newcommand{\ostar}{\mathbin{\mathpalette\make@circled\star}}
\newcommand{\make@circled}[2]{%
	\ooalign{$\m@th#1\smallbigcirc{#1}$\cr\hidewidth$\m@th#1#2$\hidewidth\cr}%
}
\newcommand{\smallbigcirc}[1]{%
	\vcenter{\hbox{\scalebox{0.77778}{$\m@th#1\bigcirc$}}}%
}
\makeatother

\begin{document}
	
	\title{Multi-View Spectral Clustering Tailored Tensor Low-Rank Representation}
	\author{Yuheng~Jia,
	Hui~Liu,
	~Junhui~Hou,~\IEEEmembership{Senior Member,~IEEE,}~
	Sam~Kwong,~\IEEEmembership{Fellow,~IEEE}, and
	Qingfu~Zhang,~\IEEEmembership{Fellow,~IEEE} 
	\thanks{Y. Jia and H. Liu are with the Department
		of Computer Science, City University of Hong Kong, Kowloon, Hong Kong,
		(e-mail: yuheng.jia@my.cityu.edu.hk; hliu99-c@my.cityu.edu.hk).} \thanks{J. Hou, S. Kwong and Q. Zhang are with the Department of Computer Science,
		City University of Hong Kong, Kowloon, Hong Kong and also with the City University of Hong Kong Shenzhen Research Institute, Shenzhen, 51800, China, (e-mail: jh.hou@cityu.edu.hk; cssamk@cityu.edu.hk; qingfu.zhang@cityu.edu.hk).}}
	\maketitle
	
	\begin{abstract}
This paper explores the problem of multi-view spectral clustering (MVSC) based on tensor low-rank modeling. Unlike the existing methods that all adopt an off-the-shelf tensor low-rank norm without considering the special characteristics of the tensor in MVSC, we design a novel structured tensor low-rank norm tailored to MVSC. Specifically, we explicitly impose a symmetric low-rank constraint and a structured sparse low-rank constraint on the frontal and horizontal slices of the tensor to characterize the intra-view and inter-view relationships, respectively. Moreover, the two constraints could be jointly  optimized  to achieve mutual refinement. On basis of the novel tensor low-rank norm, we formulate MVSC as a convex low-rank tensor recovery problem, which is   then  efficiently solved with  an augmented Lagrange multiplier based method iteratively. Extensive experimental results on five  benchmark datasets show that the proposed method outperforms state-of-the-art methods to a significant extent. Impressively, our method is able to produce perfect clustering. In addition, the parameters of our method can be easily tuned, and the proposed model is robust to different datasets, demonstrating its potential in practice. 
\end{abstract}

\begin{IEEEkeywords}
Multi-view spectral clustering, tensor low-rank representation, tensor low-rank norm. 
\end{IEEEkeywords}
	\section{Introduction}
As a promising tool to analyze data, spectral clustering  (SC) \cite{vonLuxburg2007}, which exploits the pairwise
relationship between samples, was  initially designed for the single-view data.
However, many real-world datasets may be  collected via multi-modalities  or diverse feature extractors \cite{8741173}. For example,  a self-driving car has a series of sensors to sense  the road conditions. 
An image can be represented by different types of features, e.g., texture, color, and edge. 
Those multi-view features can provide enormous information about the dataset, and the features from different views 
depict the dataset from different perspectives that may be complementary to each other.
Thus, many multi-view  spectral clustering  (MVSC) methods were proposed \cite{8501973} for processing such multi-view data. 
For example, Xia \textit{et al.} \cite{AAAIxia2014robust} first constructed a set of similarity matrices from multiple views, and then decomposed those matrices into a shared similarity matrix for the final clustering task and a series of sparse error matrices.  Cao \textit{et al.} \cite{cao2015diversity} proposed to explore the complementary information by a novel diversity-induced term. Zhao \textit{et al.} \cite{zhao2017multiAAAINMF} realized MVSC with a non-negative matrix factorization approach. 
Zhang \textit{et al.} \cite{zhang2017latent} learned a latent representation for multiple features, and simultaneously performed data reconstruction based on the learned latent representation. 
Wang \textit{et al.} \cite{wang2017exclusivity} explored the complementarity as well as the consistency among different views at the same time.
See  \cite{8336846,8471216,zhao2017multi} for the comprehensive surveys on MVSC.

Recently, by stacking the similarity matrices from multiple views as a 3-dimensional (3-D) tensor,  tensor based MVSC methods \cite{8421595} have received considerable  attention. Specifically, by exploiting the low-rankness of the tensor, this kind of  methods can automatically capture the higher-order correlations beneath the multiple features. For example, Zhang \textit{et al.} \cite{zhang2015low} proposed a low-rank tensor constrained self-representation model for MVSC, where the tensor low-rank norm was proposed by  \cite{Tensor-Low-Rank-2013-TPAMI}.
Xie \textit{et al.} \cite{IJCVxie2018unifying} extended the work of \cite{zhang2015low} with a different tensor low-rank norm defined by tensor singular value decomposition (SVD) \cite{SVD-tensor}. More importantly, the tensor is rotated  to ensure the consensus among multiple views, which improves clustering performance to a new level,   compared with the previous methods. Wu \textit{et al.}
\cite{TIPwu2019essential} proposed to learn a low-rank tensor for SC directly from a set of multiple similarity matrices.  Qu \textit{et al.} 
\cite{qu2017robust} kernelized original data to explore the non-liner relationships  among samples under the tensor MVSC framework \cite{IJCVxie2018unifying}. 
Tensor based MVSC methods were also extended to solve the time series clustering problem \cite{8812928}.

\begin{figure*}
	\begin{center}
		\includegraphics[width=\textwidth]{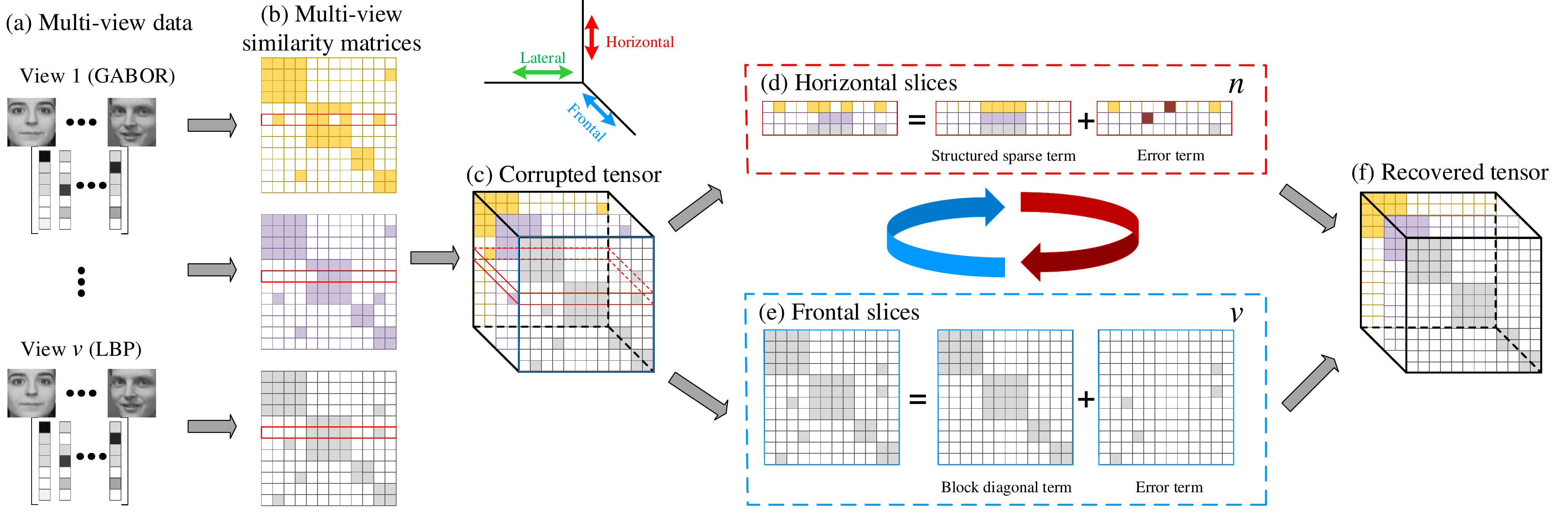}
	\end{center}
	\caption{Visual illustration of the proposed model.
	Given the input data with $v$ views (a), $v$ similarity matrices of each corresponding to a view are first constructed with a typical method (b). Then, the resulting similarity matrices are stacked to form a corrupted tensor (c). Taking the characteristics of multi-view clustering into consideration, the slices of the tensor along different dimensions own different special structures. Specifically, for the frontal slices that correspond the similarity relationships within a view,  they exhibit a block-diagonal appearance (e).  For the horizontal slices that correspond to sample-level relationships across different views, they should be structured sparse matrices, i.e., column-wise sparse (all the entries of some columns are zero) (d). We impose  a symmetric low-rank term and a structured sparse low-rank term on the frontal and horizontal slices, respectively, to seek the ideal appearances.  More importantly, the two types of low-rank representations are performed simultaneously to mutually boost each other. Finally, an enhanced low-rank tensor representation tailored to  MVSC can be achieved  (f).}
	\label{fig:short}
\end{figure*}

Although the above-mentioned tensor based MVSC methods have improved clustering performance to some extent, all of them adopt an existing tensor low-rank norm (TLRN), which is designed for general purposes without taking the unique characteristics of MVSC into account. 
That is, the existing TLRNs describe the relationships among entries of a tensor from a global perspective; however, the tensor constructed by multiple similarity matrices in MVSC has some unique local structures. 
To this end, we propose a novel TLRN tailed to MVSC, and then formulate the MVSC as a low-rank tensor learning problem. Specifically, as shown in Fig. \ref{fig:short}, ideally, the frontal slices of the tensor that capture the intra-view relationships among samples should be block-diagonal matrices, while the horizontal and lateral slices of the tensor, which capture the inter-view relationships of a typical sample from diverse views, should be column/row-wise sparse matrices. 
To pursue such unique characteristics, we propose a novel TLRN tailored to MVSC, which explicitly imposes a symmetric low-rank constraint  and a column-wise sparse low rank constraint to the frontal and horizontal slices\footnote{When the frontal slices are symmetric matrices, the horizontal slices will be the same as the transpose of the lateral slices.} of the tensor, respectively. 
Moreover, these two constraints  are implemented simultaneously to achieve mutual enhancement. Based on the MVSC tailored TLRN, we formulate MVSC as a convex low-rank tensor recovery  problem, which can be efficiently solved with  an augmented Lagrange multiplier method. We validate the proposed model on five commonly-used multi-view clustering datasets, and { experimental results show that the proposed model can dramatically improve state-of-the-art methods. More impressively, our model is able to produce perfect or nearly perfect clustering. }
We believe our perspective on MVSC will inspire this community. 

The rest of this paper is organized as follows. In Section II, we briefly  review existing TLRNs and related MVSC methods.  Section III presents the proposed 
TLRN and the associated model for MVSC as well as its numerical
solution. 
In Section IV, we validate the advantages of the proposed method with extensive experiments.  Finally, Section V concludes this paper and discusses several  future works. 
	\section{Related Work}
\subsection{Tensor Low-Rank Norm (TLRN)}
\label{tensorLRN}
As a generalization of matrices, tensors are higher order multidimensional arrays. Rank is one of the most fundamental characteristics of a matrix. Unlike that of a  matrix, the rank of a tensor is, however, not unique, and different tensor ranks were induced by different kinds of tensor decompositions \cite{8651300,8989951}. 
For example, the CANDECOMP/PARAFAC (CP) rank was induced by the CP decomposition \cite{2020-TCSVT-tensor}, which factorizes a tensor as a linear combination of rank-one tensors. 
However, estimating the CP rank of a tensor is an NP-hard problem \cite{hillar2013most}. 
As an alternative, the Tucker rank is a vector \cite{Tensor-Low-Rank-2013-TPAMI}, the $i$-th element of which is the rank of the mode-$i$ matricization of the tensor.
Liu \textit{et al.} \cite{Tensor-Low-Rank-2013-TPAMI} proposed using the sum of nuclear norms (SNN) to relax the discrete and non-convex Tucker rank. Specifically, taking a 3-D tensor $\mathcal{A}\in\mathbb{R}^{n_1\times n_2\times n_3}$ as an example, SNN is defined as
\begin{equation}
    {\rm SNN}(\mathcal{A})=\sum_{m=1}^3\zeta_m\|\mathbf{A}_m\|_*, 
\end{equation}
where $\mathbf{A}_m\in\mathbb{R}^{n_m\times (\frac{n_1n_2n_3}{n_m})}$ is a matrix by unfolding $\mathcal{A}$ along the $m$-th mode ($m=1$, $2$, and $3$), $\zeta_m$ is the weight corresponding to the $m$-th mode, and $\|\cdot\|_*$ returns the nuclear norm of a matrix. 

Recently, the tensor tubal rank was induced based on tensor SVD (t-SVD) \cite{SVD-tensor}. Specifically, t-SVD is represented as  
\begin{equation}
    \mathcal{A}=\mathcal{U}\star\mathcal{S}\star\mathcal{V},
\end{equation}
where $\mathcal{U}\in\mathbb{R}^{n_1\times n_1\times n_3}$ and $\mathcal{V}\in\mathbb{R}^{n_2\times n_2\times n_3}$ are orthogonal tensors, $\mathcal{S}\in\mathbb{R}^{n_1\times n_2\times n_3}$ is an F-diagonal tensor, 
and  
$\star$ denotes the tensor-to-tensor product. The detailed definitions of those tensor operators can be found in \cite{8606166}. According to t-SVD, the tubal rank is defined as the number of non-zero tubes in  $\mathcal{S}$, i.e.,
\begin{equation}
    {\rm tubal~rank}(\mathcal{A})=\#\{i,\mathcal{S}(i,i,:)\neq0\}.
\end{equation}
Then, Zhang \textit{et. al} \cite{zhang2014novel} proposed a tensor nuclear  norm using the t-SVD defined as 
\begin{equation}
\|\mathcal{A}\|_{\rm tSVD}= \sum_{i=1}^{\min(n_1,n_2)}\sum_{k=1}^{n_3}|\mathcal{S}(i,i,k)|,
\end{equation} 
and Lu \textit{et al.} \cite{8606166} also proposed a
novel tensor nuclear  norm based on t-SVD, but only used the first frontal slice of $\mathcal{S}$, 
i.e., 
\begin{equation}
\|\mathcal{A}\|_{\rm tSVD_{ff}}=\sum_i\mathcal{S}(i,i,1). 
\label{tSVD-Lu}
\end{equation}
Such a definition makes it consistent with the  matrix nuclear norm \cite{8606166}. 

      
\subsection{Mulit-View Spectral Clustering (MVSC)}
According to the involvement of TLRNs, we roughly divide existing MVSC methods into two categories, i.e., the non-tensor methods and tensor based methods. 

For the first category, one commonly adopted strategy is to exploit the consensus among all the views \cite{nie2018multiview,zhao2017multiAAAINMF}. For example, Kumar \textit{et al.} 
\cite{kumar2011co} proposed to learn a common representation from all the views. Xia \textit{et al.} \cite{AAAIxia2014robust} recovered a shared low-rank matrix from multiple input similarity matrices as the final affinity matrix for clustering. 
Since different views may contain valuable individual information, exploring the complementary information among them is also important. 
For example, a co-training method was proposed to alternatively use the clustering result from one view to guide the other \cite{kumar2011coICML}. 
Cao \textit{et al.} \cite{cao2015diversity} used the Hilbert Schmidt Independence Criterion as a diversity term to exploit complementary of multiple views.  
%
%
%
Moreover, some works tired to simultaneously use both the consistency and complementary information among different views \cite{8712567,wang2017exclusivity} to further improve the clustering performance. 
Besides, MVSC was also exploited in the latent space \cite{8502831,chen2020multi}.

By concatenating the similarity matrices from multiple views to form a 3-D tensor, many tensor based MVSC methods were proposed \cite{zhang2015low}. 
This kind of methods is appealing since by exploring the low-rankness of the formed tensor, the original similarity matrices can be complemented and refined from a higher-order perspective, such that both the consistency and complementarity between different views are taken into consideration naturally. 
Zhang et al. \cite{zhang2015low} proposed the  first tensor based MVSC method, which is formulated as
\begin{equation}
\begin{split}
&\min_{\mathbf{Z}^{i},\mathbf{E}^i}~\rm{SNN}(\mathcal{Z})+\lambda \|\mathbf{E}\|_{2,1}\\
&{\rm s.t. }~\mathbf{X}^{i}=\mathbf{X}^{i}\mathbf{Z}^{i}+\mathbf{E}^{i}, i=1,2,\cdots, v, \\
&\mathcal{Z}=\Phi(\mathbf{Z}^{1}, \mathbf{Z}^{2},\cdots, \mathbf{Z}^{v}),\\
&\mathbf{E}=[\mathbf{E}^{1}; \mathbf{E}^{2};\cdots; \mathbf{E}^{v}],
\end{split}
\label{ICCV-2015-SNN}
\end{equation}
where $\{\mathbf{X}^i\}\in\mathbb{R}^{d_i\times n}, i=1,2,\cdots, v$ are the data matrices from different views with $d_i$, $n$, and $v$ being the dimension of the features of the $i$th view, the number of samples, and the number of views, respectively,
$\{\mathbf{Z}^i\}\in\mathbb{R}^{n\times n}, i=1,2,\cdots, v$ are the learned similarity matrices from different views,  $\{\mathbf{E}^i\}\in\mathbb{R}^{d_i\times n}, i=1,2,\cdots, v$ denote the set of  error matrices of all views, $\Phi(\cdot)$ is a function merging  the similarity matrices $\{\mathbf{Z}^i\}$ to form  a $3$-D tensor $\mathcal{Z}\in\mathbb{R}^{n\times n \times v}$,  $\mathbf{E}\in\mathbb{R}^{\sum_{i}d_i\times n}$ concatenates all the error matrices $\{\mathbf{E}^i\}$  together along the column,  $\|\cdot\|_{2,1}$ is the column-wise $\ell_{2,1}$ norm of matrix\footnote{For a matrix $\mathbf{M}\in\mathbb{R}^{m\times n }$, $\|\mathbf{M}\|_{2,1}=\sum_{i=1}^n\|\mathbf{M}(:,i)\|_2$.}, and $\lambda>0$ is the hyper-parameter to balance the error term and the low-rank term.
By solving Eq. \eqref{ICCV-2015-SNN}, the SNN is minimized to characterize the view-to-view relationship. 
Then, Xie \textit{et al.} \cite{IJCVxie2018unifying} extended Eq. \eqref{ICCV-2015-SNN} by using  t-SVD, i.e., 
\begin{equation}
\begin{split}
&\min_{\mathbf{Z}^{i},\mathbf{E}^i}~\|\mathcal{Z}\|_{\rm tSVD}+\lambda \|\mathbf{E}\|_{2,1}\\
&{\rm s.t. }~\mathbf{X}^{i}=\mathbf{X}^{i}\mathbf{Z}^{i}+\mathbf{E}^{i}, i=1,2,\cdots, v, \\
&\mathcal{Z}=\Phi(\mathbf{Z}^1, \mathbf{Z}^{2},\cdots, \mathbf{Z}^{v})\\
&\mathbf{E}=[\mathbf{E}^{1}; \mathbf{E}^{2};\cdots; \mathbf{E}^{v}].
\end{split}
\label{-2015-SNN}
\end{equation}
It is noteworthy that \cite{IJCVxie2018unifying} rotates the tensor $\mathcal{Z}$ to better exploit the view-to-view relationship. 
With the same TLRN as  \cite{IJCVxie2018unifying}, Wu \textit{et al.} \cite{TIPwu2019essential} realized MVSC with a robust tensor  principal component analysis approach, i.e.,  
\begin{equation}
\begin{split}
&\min_{\mathcal{Z},\mathcal{E}}\|\mathcal{Z}\|_{\rm tSVD}+\lambda\|\mathcal{E}\|_1\\
&{\rm s.t. }\mathcal{W}=\mathcal{Z}+\mathcal{E},
\end{split}
\end{equation}
where $\mathcal{W}\in\mathbb{R}^{n\times n\times v}$ is the input tensor constructed by the similarity  matrices from different views,  $\mathcal{E}\in\mathbb{R}^{n\times n\times v}$ is the error tensor, and $\|\mathcal{E}\|_1=\sum_{ijk}|\mathcal{E}(i,j,k)|$ denotes the $\ell_1$ norm of a tensor. 
Yin \textit{et al.} \cite{8421595} used the  TLRN in \cite{8606166} to achieve MVSC, i.e., 
\begin{equation}
\begin{split}
\min_{\mathcal{\mathcal{Z}}}\|\mathcal{Z}\|_{\rm tSVD_{ff}}+&\lambda\|\mathcal{X}-\mathcal{X}\star\mathcal{Z}\|_F^2+\alpha\|\mathcal{Z}\|_1\\
&+\beta\sum_{i\neq j}\|\mathcal{Z}(:,:,i)-\mathcal{Z}(:,:,j)\|_F^2,
\end{split}
\label{yin-t}
\end{equation}
where  $\mathcal{X}$ is the input sample tensor, $\|\cdot\|_F$ denotes the  Frobenius norm of a tensor, and  $\lambda,\alpha,\beta$ are the hyper-parameters.
Note that in Eq. \eqref{yin-t} the  tensor-to-tensor product is introduced to better explore the higher-order information among samples, and the fourth term  enhances  the consistency among different views. 
Besides, tensor based MVSC was extended in  a semi-supervised manner by utilizing  some prior information \cite{zhang2020tensorized,9070164}. 

The above-mentioned works have validated the effectiveness of exploring the low-rankness of the tensor encoding similarity relationships in MVSC. {However, all of them adopt an existing TLRN without considering the special characteristics of MVSC, which may limit their performance in clustering.}


	\section{Proposed Method}
Given a multi-view dataset with $n$ samples and $v$ views, for each view we could build  a similarity matrix\footnote{
	Commonly used similarity matrix constrction methods could be found at \cite{vonLuxburg2007}.}
to describe the pairwise relationships between samples, i.e., $\mathbf{W}^{i}\in\mathbb{R}^{n\times n}$, where $i\in\{1,2,\cdots,v\}$ is the view index.  The tensor based MVSC methods stack the similarity matrices of all views to form a 3-D tensor denoted as  $\mathcal{W}\in\mathbb{R}^{n\times n\times v}$, where the $i$-th frontal slice of $\mathcal{W}$ is $\mathbf{W}^{i}$.

As aforementioned, the idea underlying tensor based MVSC methods is to recover a low-rank tensor $\mathcal{L}\in\mathbb{R}^{n\times n\times v}$ such that  the information
across diverse views and within each view is fully exploited from $\mathcal{W}$. 
To achieve the goal, all of the  previous methods adopt a typical TLRN which is  designed for general purposes by exploiting a tensor globally.  However, the tensor constructed by multi-view similarity matrices has  some special characteristics that need to be depicted locally, making
the existing TLRNs not be  the best choice for MVSC. 
To this end, we propose a novel TLRN tailored to the  MVSC task, which is capable of better characterizing  the special structures  of the tensor constructed by multiple similarity matrices.

\subsection{Proposed MVSC Tailored TLRN}

As can be seen  from Fig. \ref{fig:short}-(e), each frontal slice of the tensor corresponds to the intra-view similarity relationships of all the samples.  In the ideal case, it should be a block-diagonal matrix,  where only the samples from the same clusters are connected. 
To capture such a structure, we use symmetry and low-rankness to regularize the frontal slices and define
\begin{equation}
\begin{split}
&\|\mathcal{L}\|_{\small \textcircled{f}}:=\sum_{i=1}^{v}{\rm rank}(\mathbf{L}_f^{i}), \\
&~{\rm s.t.~}\mathbf{L}_f^{i}=\mathbf{L}_f^{i^\mathsf{T}},
\end{split}
\label{norm:fron}
\end{equation}
where $\mathbf{L}_f^{i}\in\mathbb{R}^{n\times n}$ is  the $i$-th frontal slice of tensor  $\mathcal{L}$ used  in MVSC, and  $\cdot^\mathsf{T}$ and ${\rm rank}(\cdot)$ denote the transpose and rank of a matrix, respectively. 

Similarly, each row of the horizontal slice (or column of the lateral slice) of the tensor depicts the pairwise relationships between a typical sample and all the other samples, i.e., the horizontal slice represents the inter-view relationships of the sample. 
As shown in Fig. \ref{fig:short}-(d), in the ideal case, it is a structured column-wise sparse matrix, where only a few columns are non-zero. Thus, we propose the following regularizer to model such a behavior, defined as 
\begin{equation}
\|\mathcal{L}\|_{\small \textcircled{h}}:=\sum_{j=1}^{n}{\rm rank}(\mathbf{L}_h^{j})+\alpha\|\mathbf{L}_h^{j}\|_{2,0},
\end{equation}
where $\mathbf{L}_h^{j}\in\mathbb{R}^{c\times n}$ denotes the $j$-th horizontal slice of $\mathcal{L}$, $\|\cdot\|_{2,0}$ is a column-wise sparse norm, i.e., counting the number of columns that are non-zero, 
and $\alpha>0$ balances the importance of the two terms. Since we restrict the symmetry of the frontal slices in Eq. \eqref{norm:fron} and the constraints on the horizontal slices equal  to the constraints on the transpose of the lateral slices, it is not necessary to add  the constraints on the lateral slices.

To pursue the ideal appearance of the tensor in  MVSC, we leverage the characteristics of both the intra-view and inter-view of the tensor, leading to an  MVSC tailored norm:
\begin{equation}
\mathcal{L}_{\oplus}:=\omega_1\|\mathcal{L}\|_{\small \textcircled{f}}+\omega_2\|\mathcal{L}\|_{\small \textcircled{h}}, 
\label{norm:prelm}
\end{equation}
where $0\leq\omega_1\leq1$ and $\omega_2=1-\omega_1$  balance the contributions of $\|\mathcal{L}\|_{\small \textcircled{f}}$ and $\|\mathcal{L}\|_{\small \textcircled{h}}$. By minimizing Eq. \eqref{norm:prelm}, both the intra-view and inter-view relationships are optimized at the same time to boost each other.

However, in Eq.\eqref{norm:prelm}, the discrete and non-smooth properties of both  the $\rm rank(\cdot)$ function and the $\ell_{2,0}$ norm make it inefficient to solve. We thus relax them by their convex hulls, i.e., the unclear norm $\|\cdot\|_*$ and the  $\ell_{2,1}$ norm, respectively.
Finally, the proposed convex MVSC tailored TLRN is written as 
\begin{equation}
\begin{split}
    &\mathcal{L}_{\ostar}:=\omega_1\sum_{i=1}^v\|\mathbf{L}_f^i\|_*+\omega_2\left(\sum_{j=1}^n\|\mathbf{L}_h^j\|_*+\alpha\|\mathbf{L}_h^j\|_{2,1}\right),\\
    &{\rm s.t.,}~\mathbf{L}_f^i=\mathbf{L}_f^{i^\mathsf{T}},\forall i\in\{1,2,\cdots,v\}.
\end{split}
\label{norm:final}
\end{equation}
Since $\mathcal{L}_{\ostar}$ is the linear combination of two kinds of valid norms (nuclear norm and $\ell_{2,1}$ norm), the proposed MVSC tailored TLRN is also a valid norm. 




\subsection{Proposed MVSC Model}
Based on the proposed TLRN in Eq. \eqref{norm:final}, 
we cast the MVSC as a low-rank tensor recovery problem, which is expressed as 
\begin{equation}
\begin{split}
&
\min_{\mathcal{L},\mathcal{E}}\|\mathcal{L}\|_{\ostar}+\lambda\|\mathcal{E}\|_{F}^2\\
&~{\rm s.t.,}\mathcal{W}=\mathcal{L}+\mathcal{E},
\end{split}
\label{model}
\end{equation}
where  $\mathcal{W}$ is an observed tensor with corruptions, $\mathcal{E}\in\mathbb{R}^{n\times n \times v}$ is the noise error tensor, 
$\|\cdot\|_F$ extends the matrix Frobenius norm to a tensor case, i.e., $\|\mathcal{E}\|_F=\sqrt{\sum_{ijk}\mathcal{E}_{ijk}^2}$, and $\lambda$ is a positive penalty parameter. By optimizing Eq. \eqref{model}, the recovered tensor $\mathcal{L}$ will be encouraged to conform to the ideal appearance of a tensor constructed by the ideal similarity matrices. 
Then, the high clustering performance on $\mathcal{L}$ can be expected.






Note that unlike the well-known  robust principal component analysis (RPCA) \cite{10.1145/1970392.1970395} and tensor RPCA \cite{8606166}, which impose an  $\ell_1$ norm on the error matrix/tensor, our model adopts  a Frobenius-type norm to regularize  the error tensor. The reason is that a sparse regularizer on $\mathcal{E}$ will conflict with the $\ell_{2,1}$ norm on the horizontal slices. 
The subsequent experimental results also validate the superiority of the proposed model.

\textit{Differences between Ours and  SNN \cite{Tensor-Low-Rank-2013-TPAMI}.}
Although both our TLRN and SNN are the sum of matrix nuclear norms, they are essentially different. 
Specifically, the involved matrices in SNN are the matricization of a tensor along different modes. While, the matrices in our TLRN are the frontal and  horizontal slices of a tensor.  
In addition,  we impose a symmetric constraint and a  column-wise sparse constraint on the frontal and  horizontal slices, respectively. 
Such a different mathematic expression reveals the  essential difference between them, i.e.,  the proposed TLRN is tailored to MVSC by seeking the ideal appearance for the tensor constructed by multiple similarity matrices,  
while SNN is just a relaxation of the tensor Tucker rank for general low-rankness. 
As will be shown in Section IV-B, the significant superiority of our method in terms of clustering performance over SNN and other TLRNs substantiates that the proposed TLRN is indeed  tailored to MVSC. 


\subsection{Optimization Algorithm}
We use the inexact augmented Lagrange multiplier (IALM)  \cite{lin2010augmented} method to solve the resulting convex model in Eq. \eqref{model}. Specifically, the IALM converts the original problem into several smaller sub-problems, each of which is relatively easier to solve. By introducing three auxiliary tensors, i.e., 
$\mathcal{L}_1=\mathcal{L}_2=\mathcal{L}_3=\mathcal{L}$, Eq. \eqref{model}  is equivalently rewritten as 
\begin{small}
\begin{equation}
\begin{split}
&\min_{\substack{\mathcal{L}_{1},\mathcal{L}_2,\\ \mathcal{L}_3,\mathcal{L},\mathcal{E}}}\omega_1\sum_{i=1}^{v}\|\mathbf{L}_{1_f}^i\|_*+\omega_2\sum_{j=1}^{n}\left(\|\mathbf{L}_{2_h}^j\|_*+\alpha\|\mathbf{L}_{3_h}^j\|_{2,1}\right)+\lambda\|\mathcal{E}\|_F^2\\
&{\rm s.t.}~\mathcal{W}=\mathcal{L}_k+\mathcal{E}, \mathcal{L}_k=\mathcal{L},\forall k\in\{1,2,3\}, \mathbf{L}_{1}^i=\mathbf{L}_{1}^{i^\mathsf{T}},\forall i\{1,\cdots,v\}.
\end{split}
\label{model:equal}
\end{equation}\end{small}
The augmented Lagrange form of Eq. \eqref{model:equal} is 
\begin{small}
\begin{equation}
\begin{split}
&\argmin_{\substack{\mathcal{L}_1,\mathcal{L}_2,\\ \mathcal{L}_3,\mathcal{L},\mathcal{E}}}\omega_1\sum_{i=1}^{v}\|\mathbf{L}_{1_f}^i\|_*+\omega_2\sum_{j=1}^{n}\left(\|\mathbf{L}_{2_h}^j\|_*+\alpha\|\mathbf{L}_{3_h}^j\|_{2,1}\right)
+\lambda\|\mathcal{E}\|_F^2\\
&+\sum_{k=1}^{3}\left(\frac{\mu}{2}\left\|\mathcal{W}-\mathcal{L}_k-\mathcal{E}+\frac{\mathcal{Y}_{1k}}{\mu}\right\|_F^2+\frac{\mu}{2}\left\|\mathcal{L}-\mathcal{L}_k+\frac{\mathcal{Y}_{2k}}{\mu}\right\|_F^2\right)\\
&{\rm s.t.,}~\mathbf{L}_{1}^i=\mathbf{L}_{1}^{i^\mathsf{T}},\forall i\in\{1,\dots,v\},
\end{split}
\label{alm}
\end{equation}\end{small}where $\mu>0$ introduces the Lagrange multiplier terms, and $\mathcal{Y}_{1k}\in\mathbb{R}^{n\times n\times v},k\in\{1,2,3\}$ and $\mathcal{Y}_{2k}\in\mathbb{R}^{n\times n\times v},k\in\{1,2,3\}$ are the Lagrange multipliers. The IALM splits Eq. \eqref{alm} into several sub-problems, and iteratively solves those sub-problems until convergence.

Specifically, the $\mathcal{E}$ sub-problem is written as
\begin{equation}
\begin{split}
&\min_{\substack{\mathcal{E}}}\lambda\|\mathcal{E}\|_F^2+\sum_{k=1}^{3}\frac{\mu}{2}\left\|\mathcal{W}-\mathcal{L}_k-\mathcal{E}+\frac{\mathcal{Y}_{1k}}{\mu}\right\|_F^2,\\
\end{split}
\end{equation}
which is a set of unconstrained quadratic equations in element-wise. Therefore, the closed-form solution is obtained as
\begin{equation}
    \mathcal{E}=\mu\sum_{k=1}^3\left(\mathcal{W}-\mathcal{L}_k+\frac{\mathcal{Y}_{1k}}{\mu}\right)/(2\lambda+3\mu).
\label{sol:E}
\end{equation}


The $\mathcal{L}$ sub-problem is written as 
\begin{equation}
\begin{split}
&\min_{\substack{\mathcal{L}}}\sum_{k=1}^{3}\frac{\mu}{2}\left\|\mathcal{L}-\mathcal{L}_k+\frac{\mathcal{Y}_{2k}}{\mu}\right\|_F^2,\\
\end{split}
\end{equation}
which is also a set of unconstrained  quadratic equations with the closed-form solution:
\begin{equation}
\mathcal{L}=\frac{1}{3}\sum_{k=1}^{3}\left(\mathcal{L}_k-\frac{\mathcal{Y}_{2k}}{\mu}\right).
\label{sol:L}
\end{equation}

For $\mathcal{L}_k,k\in\{1,2,3\}$ sub-problems, removing the irrelevant  terms, we have 
\begin{equation}
\begin{split}
&\min_{\substack{\mathcal{L}_k}}\frac{1}{2\mu}f(\mathcal{L}_k)+\frac{1}{2}\left\|\mathcal{L}_k-\frac{1}{2}\left(\mathcal{W}+\mathcal{L}-\mathcal{E}+\frac{\mathcal{Y}_{1k}+\mathcal{Y}_{2k}}{\mu}\right)\right\|_F^2\\
&{\rm s.t.,}\mathbf{L}_{1}^i=\mathbf{L}_{1}^{i^\mathsf{T}},\forall i\in\{1,\dots,v\},
\end{split}
\label{sol:Lk}
\end{equation}
where $f(\mathcal{L}_1)=\omega_1\sum_{i=1}^v\|\mathbf{L}_{1_f}^i\|_*$, $f(\mathcal{L}_2)=\omega_2\sum_{j=1}^n\|\mathbf{L}_{2_h}^j\|_*$, and $f(\mathcal{L}_3)=\omega_2\alpha\sum_{j=1}^n\|\mathbf{L}_{3_h}^j\|_{2,1}$. These  three sub-problems are symmetric unclear norm minimization, unclear norm minimization, and $\ell_{2,1}$ norm minimization problems, each of which has a closed-form solution
%
, i.e.,
\begin{equation}
\mathcal{L}_{1}(:,:,i)=\mathbf{U}_f^i \mathcal{T}\left(\mathbf{\Sigma}_f^i-\frac{\omega_1}{2\mu}\mathbf{I}_n\right) \mathbf{V}_f^i,~
i=1,2,\cdots, v,
\label{L-sub1}
\end{equation}
\begin{equation}
\mathcal{L}_{2}(:,j,:)=\mathbf{U}_h^j \mathcal{T}\left(\mathbf{\Sigma}_h^j-\frac{\omega_2}{2\mu}\mathbf{I}_v\right) \mathbf{V}_h^j,~
j=1,2,\cdots, n,
\label{L-sub2}
\end{equation}
and 
\begin{small}
\begin{equation}
\begin{split}
\mathcal{L}_{3}(:,j,p)=
&\begin{cases}
\frac{\|\mathcal{A}_{3}(:,j,p)
	\|_2-\frac{\omega_2\alpha}{2\mu}}{\|\mathcal{A}_{3}(:,j,p)
	\|_2}\mathcal{A}_{3}(:,j,p),&{\rm if}~\frac{\omega_2\alpha}{2\mu}<\|\mathcal{A}_{3}(:,j,p)
\|_2\\
0,&{\rm otherwise},\\
\end{cases}\\
&j=1,2,\cdots, n,\forall p,
\end{split}
\label{L-sub3}
\end{equation}
\end{small}where $\mathbf{U}_f^i\mathbf{\Sigma}_f^i \mathbf{V}_f^i, i=1,\cdots, v$ and  $\mathbf{U}_h^j\mathbf{\Sigma}_h^j \mathbf{V}_h^j, j=1,\cdots, n$  denote the SVD of  $\frac{1}{2}\left(\mathcal{A}_1(:,:,i)+\mathcal{A}_1(:,:,i)^\mathsf{T}\right),  i=1,2,\cdots, v$ and $\mathcal{A}_2(:,j,:),  j=1,2,\cdots, n$, respectively, $\mathcal{A}_k=\mathcal{W}+\mathcal{L}-\mathcal{E}+\frac{\mathcal{Y}_{1k}+\mathcal{Y}_{2k}}{\mu}, k\in\{1,2,3\}$, $\mathbf{I}_n$ and $\mathbf{I}_v$ are the identity   matrices of size $n\times n$ and $v\times v$, respectively, $\|\cdot\|_2$ denotes the $\ell_2$ norm of a vector, and $\mathcal{T}(\cdot)$ is a thresholding operator in element-wise, i.e.,   $\mathcal{T}(x):=\max(0,x)$.

The Lagrange multipliers  and $\mu$ are updated via
\begin{equation}
\begin{cases}
\mathcal{Y}_{1k}=\mathcal{W}-\mathcal{L}_k-\mathcal{E},\forall k\in\{1,2,3\}\\
\mathcal{Y}_{2k}=\mathcal{L}-\mathcal{L}_k,\forall k\in\{1,2,3\}\\
\mu={\rm min}(1.1*\mu,\mu_{max}),
\end{cases}
\label{sol:Y}
\end{equation}
where $\mu_{max}$ gives an upper bound for $\mu$. Finally, the overall optimization procedure is summarized in Aglorithm \ref{alg: dp}. 
\begin{algorithm}[!t]
	\caption{Optimization Solution to Eq. \eqref{model}}
	\begin{algorithmic}[1]
		\renewcommand{\algorithmicrequire}{\textbf{Input:}}
		\renewcommand{\algorithmicensure}{\textbf{Initialize:}}
		\REQUIRE Similarity matrix sets $\{\mathbf{W}^i|_{i=1}^v\}$, $\omega_1$, $\lambda$, $\alpha$;
		\ENSURE $\omega_2=1-\omega_1$, $\mathcal{L}=\mathcal{L}_{k}=\mathcal{Y}_{1k}=\mathcal{Y}_{2k}=\mathcal{O}^{n\times n \times v},\forall k,\mu=10^{-4},\mu_{\rm max}=10^{8}$, where $\mathcal{O}$ denotes a tensor with all entries equal to zeros;
		\STATE Form a tensor $\mathcal{W}$ by stacking $\{\mathbf{W}^i|_{i=1}^v\}$;
		\WHILE{not converged}
		\STATE Update $\mathcal{L}_{k}, k\in\{1,2,3\}$ by Eqs. \eqref{L-sub1}-\eqref{L-sub3};%
		\STATE Update $\mathcal{L}$ by Eq. \eqref{sol:L};%
		\STATE Update $\mathcal{E}$ by Eq. \eqref{sol:E};
		\STATE Update $\mathcal{Y}_{1k},\mathcal{Y}_{2k},k\in\{1,2,3\}$ and $\mu$ by  Eq. \eqref{sol:Y};
		\ENDWHILE
		\STATE \textbf{Output:} $\mathcal{L}$.
	\end{algorithmic}
	\label{alg: dp}
\end{algorithm}

After solving  Eq. \eqref{model}, we could construct the final similarity matrix by averaging  $\mathcal{L}$ along the frontal direction, i.e.,
$\mathbf{S}=\frac{1}{v}\sum_{i}^v\mathbf{L}_f^i$
and perform SC on $\mathbf{S}$ to generate the clustering result. 
\subsection{Computational Complexity Analysis}
The computational complexity of Algorithm 1 is dominated by step 3 which involves two nuclear norm minimization problems regarding $\mathcal{L}_1$ and $\mathcal{L}_2$ and one $\ell_{21}$ norm minimization problem regarding $\mathcal{L}_3$.
The $\mathcal{L}_1$ sub-problem needs to solve $v$ SVDs of $n\times n$ matrices, which leads to a computational complexity of $\mathsf{O}(vcn^2)$ with partial SVD \cite{lin2011linearized}, where $c$ is the number of clusters. 
The $\mathcal{L}_2$ sub-problem involves $n$ SVDs of $v\times n$ matrices resulting in a total complexity of $\mathsf{O}(v^2n^2)$. 
The complexity of $\mathcal{L}_3$ sub-problem is much lower than those of the $\mathcal{L}_1$ and $\mathcal{L}_2$ sub-problems. 
Therefore, the total computational complexity  of Algorithm 1 is $\mathsf{O}(v^2n^2+cvn^2)$.

	\begin{figure}[!t]
	
	\begin{minipage}[b]{0.48\linewidth}
		\centering
		\centerline{\epsfig{figure=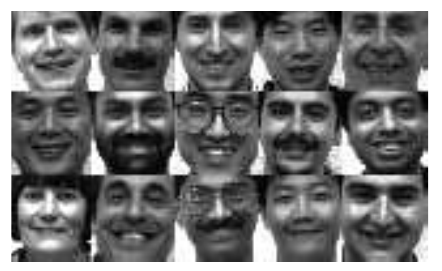,width=4cm}}
		\centerline{(a) Yale}\medskip
	\end{minipage}
	\begin{minipage}[b]{0.48\linewidth}
		\centering
		\centerline{\epsfig{figure=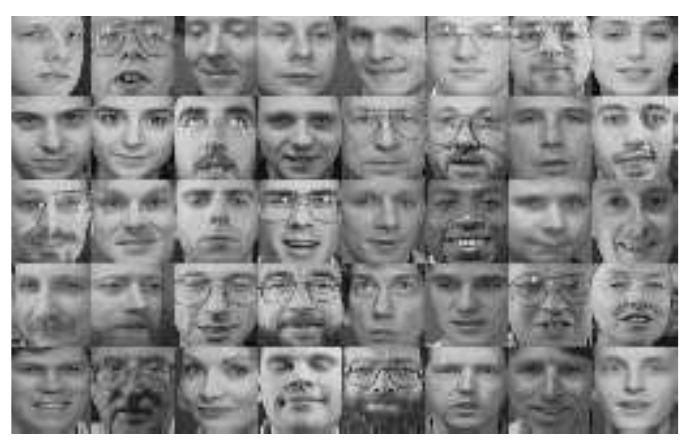,width=4cm}}
		\centerline{(b) ORL}\medskip
	\end{minipage}\\
	\begin{minipage}[b]{0.52\linewidth}
		\centering
		\centerline{\epsfig{figure=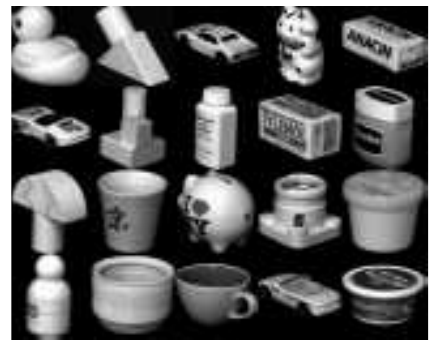,width=4.5cm}}
		\centerline{(c) Coil20}\medskip
	\end{minipage}
	\begin{minipage}[b]{0.47\linewidth}
		\centering
		\centerline{\epsfig{figure=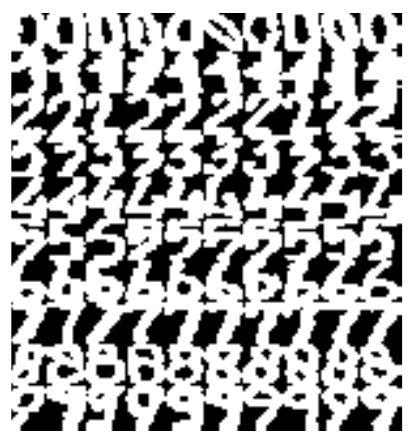,width=3.6cm}}
		\centerline{(d) UCI-digital}\medskip
	\end{minipage}
	\\
	\begin{minipage}[b]{1\linewidth}
		\centering
		\centerline{\epsfig{figure=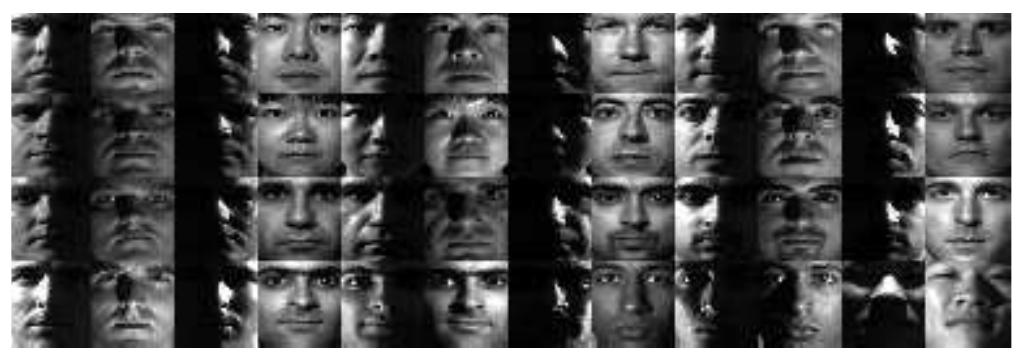,width=8.5cm}}
		\vspace{-0.1cm}
		\centerline{(e) YaleB }\medskip
	\end{minipage}\\
	\caption{Sample images from five datasets used in the experiments. (a)
		Yale, (b) ORL, (c) Coil20, (d) UCI-digital, and (e) YaleB.}
	\label{fig:samples}
\end{figure}
\begin{table}  
    \caption{Dataset Summary}
    \begin{center}
		\begin{tabular}{cccccc}
		\hline\hline
		Dataset  & Samples & Views & Clusters & Type\\
		\hline
		Coil20  & 1440 & 3 & 20 & Object \\
		Yale  & 165 & 3 & 15 & Face \\
		YaleB  & 640 & 3 & 10 & Face \\
		ORL  & 400 & 3 & 40 & Face \\
		UCI-digit  & 2000 & 3 & 10 & Digit \\
		\hline\hline
	\end{tabular}
	\label{tab:dataSet}
	\end{center}
\end{table}

\begin{table*}
\caption{Clustering results (mean$\pm$std) on \textbf{Coil20}. We set $\omega_1=0.5$, $\alpha=5$, and $\lambda=15$ in the proposed method. 
			The highest and the second highest values under each metric are \textbf{bolded} and \underline{underlined}, respectively.  
			}
	\centering
		\begin{tabular}{cccccccc}
			\hline\hline
			Method  & ACC & NMI & ARI & F1-score & Precision & Recall & Purity \\
			\hline
			L-MSC  \cite{8502831}
			&$0.736\pm0.017$
			&$0.807\pm0.013$
			&$0.661\pm0.022$
			&$0.67\pm0.021$
			&$0.64\pm0.024$
			&$0.715\pm0.017$	
			&$0.767\pm0.012$
			\\
			DiMSC \cite{cao2015diversity}
			&	$0.694\pm0.015$
			&	$0.728\pm0.010$
			&	$0.587\pm0.015$
			&	$0.608\pm0.014$
			&	$0.589\pm0.016$
			&	$0.629\pm0.013$
			&	$0.701\pm0.015$
			\\
			
			MCLES \cite{chen2020multi}
			&   $0.706\pm0.026$	
			&   $0.740\pm0.019$
			&   $0.521\pm0.032$
			&   $0.553\pm0.029$	
			&   $0.505\pm0.036$	
			&   $0.611\pm0.021$	
			&   $0.709\pm0.026$\\
			AWP \cite{nie2018multiview}
			&   $0.896\pm0.000	$
			&   $0.968\pm0.000	$
			&   $0.892\pm0.000	$
			&   $0.897\pm0.000	$
			&   $0.847\pm0.000	$
			&   $0.954\pm0.000	$
			&   $0.915\pm0.000$\\ \cdashline{1-8}
			LT-MSC  \cite{zhang2015low}  
			&$0.770\pm0.013$
			&$0.873\pm0.005$
			&$0.725\pm0.018$
			&$0.740\pm0.017$
			&$0.696\pm0.027$
			&$0.790\pm0.005$
			&$0.817\pm0.005$
			
			\\
			Ut-SVD-MSC    \cite{IJCVxie2018unifying}   
			&	$   0.794\pm0.015$
			&	$	0.877\pm0.005$
			&	$	0.755\pm0.018$
			&	$	0.767\pm0.017$
			&	$	0.746\pm0.025$
			&	$	0.790\pm0.009$
			&	$	0.818\pm0.009$
			\\
			t-SVD-MSC    \cite{IJCVxie2018unifying}  
			&   $0.836\pm0.007$ 
			&	$0.924\pm0.002$	
			&   $0.799\pm0.011$
			&	$0.810\pm0.010$
			&	$0.759\pm0.021$
			&	$0.869\pm0.003$
			&	$0.879\pm0.001$\\
			ETLMC \cite{TIPwu2019essential}
			&$\underline{0.956\pm0.037}	$
			&$\underline{0.977\pm0.012}	$
			&$\underline{0.950\pm0.035}	$
			&$\underline{0.952\pm0.033}	$
			&$\underline{0.937\pm0.049}	$
			&$\underline{0.969\pm0.019}	$
			&$\underline{0.965\pm0.027}$\\
			SCMV-3DT \cite{8421595}
			&   $0.761\pm0.011$	
			&   $0.857\pm0.004$
			&   $0.707\pm0.013$
			&   $0.722\pm0.013$	
			&   $0.685\pm0.017$
			&   $0.764\pm0.010$	
			&   $0.802\pm0.007$\\
			\cdashline{1-8}
			SNN \cite{Tensor-Low-Rank-2013-TPAMI}
			&   $0.850\pm0.016$
			&   $0.927\pm0.005$	
			&   $0.819\pm0.021$
			&   $0.828\pm0.020$	
			&   $0.788\pm0.034$	
			&   $0.874\pm0.008$
			&   $0.885\pm0.007$\\
			
			TRPCA \cite{8606166}
			&   $0.853\pm0.018$	
			&   $0.906\pm0.004$	
			&   $0.818\pm0.016$	
			&   $0.827\pm0.015$	
			&   $0.805\pm0.021$	
			&   $0.851\pm0.008$	
			&   $0.882\pm0.010$\\
            \cdashline{1-8}

			Proposed       & $\mathbf{1.000\pm0.000}$	&$\mathbf{1.000\pm0.000}$	&$\mathbf{1.000\pm0.000}$ &$\mathbf{1.000\pm0.000}$	&$\mathbf{1.000\pm0.000}$	&$\mathbf{1.000\pm0.000}$	&$\mathbf{1.000\pm0.000}$\\
			\hline\hline
		\end{tabular}
		\label{tab:coil20}
	\end{table*}

\begin{table*}
\caption{Clustering results (mean$\pm$std) on \textbf{Yale}. We set $\omega_1=0.5$, $\alpha=5$, and $\lambda=15$ in the proposed method.}
		\centering
			\begin{tabular}{cccccccc}
				\hline\hline
				Method  & ACC & NMI & ARI & F1-score & Precision & Recall & Purity \\
				\hline
				
				L-MSC \cite{8502831}
				& $0.735\pm0.022$
				& $0.741\pm0.012$
				& $0.528\pm0.019$
				& $0.559\pm0.018$
				& $0.518\pm0.019$
				& $0.607\pm0.019$
				& $0.735\pm0.022$
				\\
				DiMSC  \cite{cao2015diversity}
				& $0.694\pm0.016$
				& $0.694\pm0.013$
				& $0.505\pm0.021$
				& $0.536\pm0.020$
				& $0.523\pm0.022$
				& $0.549\pm0.018$
				& $0.698\pm0.014$
				\\
				MCLES \cite{chen2020multi}
				& $0.706\pm0.026$
				& $0.740\pm0.019$
				& $0.521\pm0.032$
				& $0.553\pm0.029$
				& $0.505\pm0.036$	
				& $0.611\pm0.021$	
				& $0.708\pm0.026$\\
				AWP \cite{nie2018multiview}
				& $0.587\pm0.000	$	
				& $0.601\pm0.000	$
				& $0.309\pm0.000		$
				& $0.364\pm0.000		$
				& $0.278\pm0.000		$
				& $0.529\pm0.000		$
				& $0.587\pm0.000	$\\
				\cdashline{1-8}
				LT-MSC \cite{zhang2015low}
				& $0.739\pm0.014$
				& $0.767\pm0.009$
				& $0.597\pm0.013$
				& $0.623\pm0.012$
				& $0.601\pm0.012$
				& $0.647\pm0.014$
				& $0.739\pm0.014$
				\\
				Ut-SVD-MSC  \cite{IJCVxie2018unifying}     
				& $0.732\pm0.007$	
				& $0.762\pm0.008$
				& $0.580\pm0.016$
				& $0.607\pm0.015$
				& $0.576\pm0.019$
				& $0.641\pm0.011$
				& $0.732\pm0.007$
				\\
				t-SVD-MSC  \cite{IJCVxie2018unifying}     
				& $\underline{0.955\pm0.040}$	
				& $\underline{0.954\pm0.030}$	
				& $\underline{0.913\pm0.057}$	
				& $\underline{0.918\pm0.053}$
				& $\underline{0.909\pm0.059}$
				& $\underline{0.928\pm0.047}$
				& $\underline{0.955\pm0.040}$
				\\
				ETLMC \cite{TIPwu2019essential}
				& $0.677\pm0.074$
				& $0.725\pm0.062$
				& $0.558\pm0.089$
				& $0.585\pm0.083$
				& $0.572\pm0.085$
				& $0.599\pm0.082$
				& $0.677\pm0.074$
				\\
				
				SCMV-3DT \cite{8421595}
				& $0.711\pm0.007$	
				& $0.731\pm0.008$	
				& $0.530\pm0.013$	
				& $0.560\pm0.012$	
				& $0.536\pm0.014$
				& $0.587\pm0.011$	
				& $0.716\pm0.006$\\ \cdashline{1-8}
				SNN \cite{Tensor-Low-Rank-2013-TPAMI}
				& $0.853\pm0.043	$
				& $0.883\pm0.032$	
				& $0.743\pm0.074$	
				& $0.760\pm0.069	$
				& $0.726\pm0.081$	
				& $0.798\pm0.056	$
				& $0.853\pm0.043$\\
				TRPCA \cite{8606166}
				& $0.760\pm0.039	$
				& $0.787\pm0.034$	
				& $0.629\pm0.049$	
				& $0.652\pm0.046$	
				& $0.633\pm0.044$	
				& $0.673\pm0.050$
				& $0.767\pm0.037$\\
                \cdashline{1-8}
				Proposed       
				&   $\mathbf{0.988\pm0.001}$ 
				& 	$\mathbf{0.988\pm0.001}$ 
				&	$\mathbf{0.975\pm0.003}$
				&	$\mathbf{0.976\pm0.003}$
				&	$\mathbf{0.974\pm0.003}$
				&	$\mathbf{0.979\pm0.002}$
				&	$\mathbf{0.988\pm0.001}$\\
				\hline\hline
			\end{tabular}
			\label{tab:Yale}
		\end{table*}

\section{Experiments}
\subsection{Experiment Settings}
\begin{table*}
\caption{Clustering results (mean$\pm$std) on \textbf{YaleB}. We set $\omega_1=0.4$, $\alpha=9$, and $\lambda=15$ in the proposed method.}
	\centering
		\begin{tabular}{cccccccc}
			\hline\hline
			Method  & ACC & NMI & ARI & F1-score & Precision & Recall & Purity\\
			\hline
			L-MSC \cite{8502831}
			& $ 0.472\pm0.002$	
			& $0.438\pm0.004$
			& $0.187\pm0.002$
			& $0.283\pm0.002$
			& $0.237\pm0.002$
			& $0.353\pm0.003$
			& $0.475\pm0.002$
			\\
			DiMSC \cite{cao2015diversity}
			& $0.520\pm0.003$
			& $0.496\pm0.003$
			& $0.32\pm0.004$
			& $0.393\pm0.004$
			& $0.375\pm0.004$
			& $0.412\pm0.003$	
			& $0.523\pm0.003$
			\\
			MCLES \cite{chen2020multi}
			& $0.426\pm0.001$
			& $0.420\pm0.001$
			& $0.129\pm0.001$
			& $0.240\pm0.001$
			& $0.185\pm0.001$
			& $0.344\pm0.001$
			& $0.426\pm0.001$\\
			AWP \cite{nie2018multiview}
			& $0.514\pm0.000	$
			& $0.567\pm0.000	$
			& $0.197\pm0.000$	
			& $0.313\pm0.000$	
			& $0.213\pm0.000$	
			& $0.588\pm0.000$	
			& $0.531\pm0.000$\\
			\cdashline{1-8}
			LT-MSC \cite{zhang2015low}
			& $0.617\pm0.002$
			& $0.623\pm0.006$
			& $0.413\pm0.006$
			& $0.491\pm0.005$
			& $0.466\pm0.005$
			& $0.521\pm0.005$
			& $0.620\pm0.002$
			\\
			Ut-SVD-MSC \cite{IJCVxie2018unifying}     
			& $0.502\pm0.001$
			& $0.506\pm0.001$
			& $0.236\pm0.001$
			& $0.325\pm0.001$
			& $0.276\pm0.001$
			& $0.395\pm0.001$
			& $0.503\pm0.001$
			\\
			t-SVD-MSC  \cite{IJCVxie2018unifying}     
			& ${0.668\pm0.008}$
			& ${0.696\pm0.006}$
			& ${0.513\pm0.008}$	
			& ${0.563\pm0.007}$	
			& ${0.539\pm0.007}$	
			& ${0.590\pm0.008}$
			& ${0.669\pm0.007}$
			\\
			ETLMC \cite{TIPwu2019essential}
			& $0.325\pm0.011$
			& $0.307\pm0.021$
			& $0.179\pm0.019$
			& $0.262\pm0.017$
			& $0.257\pm0.017$
			& $0.267\pm0.017$
			& $0.332\pm0.010$
			\\
			
			SCMV-3DT \cite{8421595}
			& $0.410\pm0.001	$
			& $0.413\pm0.002$	
			& $0.185\pm0.002	$
			& $0.276\pm0.001	$
			& $0.244\pm0.002	$
			& $0.318\pm0.001	$
			& $0.413\pm0.001$
			\\ \cdashline{1-8}
			SNN \cite{Tensor-Low-Rank-2013-TPAMI}
			& $\underline{0.687\pm0.001}$	
			& $\underline{0.708\pm0.001}$	
			& $\underline{0.545\pm0.002}$	
			& $\underline{0.592\pm0.001}$
			& $\underline{0.573\pm0.002}$
			& $\underline{0.612\pm0.001}$
			& $\underline{0.687\pm0.001}$\\
			TRPCA \cite{8606166}
			& $0.682\pm0.003$
			& $0.699\pm0.004$	
			& $0.534\pm0.005$	
			& $0.581\pm0.005$	
			& $0.560\pm0.005$	
			& $0.605\pm0.004$	
			& $0.682\pm0.003$\\
            \cdashline{1-8}
			Proposed      
			& $\mathbf{0.954\pm0.002}$
			& $\mathbf{0.908\pm0.001}$
			& $\mathbf{0.900\pm0.003}$	
			& $\mathbf{0.910\pm0.002}$
			& $\mathbf{0.909\pm0.003}$	
			& $\mathbf{0.912\pm0.003}$	
			& $\mathbf{0.954\pm0.003}$\\
			\hline\hline
		\end{tabular}
		\label{tab:eYaleB}
	\end{table*}
\begin{table*}
		\caption{Clustering results (mean$\pm$std) on \textbf{ORL}. We set $\omega_1=0.5$, $\alpha=5$, and $\lambda=15$ in the proposed method.}
	\centering
		\begin{tabular}{cccccccc}
			\hline\hline
			Method  & ACC & NMI & ARI & F1-score & Precision & Recall & Purity\\
			\hline
			
			L-MSC  \cite{8502831}
			& $0.823\pm0.026$	
			& $0.930\pm0.010$	
			& $0.775\pm0.033$	
			& $0.780\pm0.032$
			& $0.728\pm0.039$	
			& $0.840\pm0.024$	
			& $0.859\pm0.019$
			\\
			DiMSC  \cite{cao2015diversity}
			& $0.817\pm0.030$	
			& $0.917\pm0.013$	
			& $0.757\pm0.036$	
			& $0.767\pm0.035$	
			& $0.727\pm0.041$	
			& $0.806\pm0.030$	
			& $0.843\pm0.025$
			\\
			MCLES \cite{chen2020multi}
			&$0.792\pm0.021$	
			&$0.908\pm0.008$	
			&$0.705\pm0.035$	
			&$0.713\pm0.034$	
			&$0.647\pm0.046$	
			&$0.795\pm0.018$	
			&$0.833\pm0.015$\\
			AWP \cite{nie2018multiview}
			&$0.632\pm0.000$	
			&$0.872\pm0.000$	
			&$0.547\pm0.000	$
			&$0.560\pm0.000$	
			&$0.429\pm0.000$	
			&$0.807\pm0.000$	
			&$0.675\pm0.000$\\
			\cdashline{1-8}
			LT-MSC \cite{zhang2015low}
			& $0.808\pm0.022$
			& $0.915\pm0.010$
			& $0.747\pm0.029$
			& $0.753\pm0.028$
			& $0.708\pm0.034$
			& $0.805\pm0.023$
			& $0.838\pm0.015$
			\\
			Ut-SVD-MSC   \cite{IJCVxie2018unifying} 
			& $0.831\pm0.017$
			& $0.934\pm0.004$
			& $0.787\pm0.015$
			& $0.792\pm0.015$
			& $0.742\pm0.023$
			& $0.850\pm0.011$
			& $0.864\pm0.012$
			\\
			t-SVD-MSC   \cite{IJCVxie2018unifying}  
			& $\underline{0.964\pm0.018}$ 
			& $\underline{0.992\pm0.003}$
			& $\underline{0.965\pm0.019}$
			& $\underline{0.965\pm0.018}$
			& $\underline{0.945\pm0.030}$
			& $\underline{0.987\pm0.006}$
			& $\underline{0.974\pm0.015}$
			\\
			ETLMC  \cite{TIPwu2019essential}
			& $0.946\pm0.018$
			& $0.986\pm0.005$
			& $0.942\pm0.020$
			& $0.943\pm0.019$
			& $0.918\pm0.026$
			& $0.970\pm0.014$
			& $0.960\pm0.014$
			\\

			SCMV-3DT \cite{8421595}
			&$0.839\pm0.012	$
			&$0.908\pm0.007	$
			&$0.763\pm0.018	$
			&$0.769\pm0.017	$
			&$0.747\pm0.020	$
			&$0.792\pm0.016	$
			&$0.852\pm0.012$\\
			\cdashline{1-8}
			SNN \cite{Tensor-Low-Rank-2013-TPAMI}
			&$0.861\pm	0.021$	
			&$0.935\pm	0.009$	
			&$0.815\pm	0.024$	
			&$0.819\pm	0.023$	
			&$0.790\pm	0.027$	
			&$0.850\pm	0.022$	
			&$0.882\pm	0.017$\\
			TRPCA \cite{8606166}
			&$0.932\pm	0.015	$
			&$0.978\pm	0.005	$
			&$0.922\pm	0.017	$
			&$0.924\pm	0.016	$
			&$0.896\pm	0.022	$
			&$0.954\pm	0.013	$
			&$0.947\pm	0.011$\\
			\cdashline{1-8}
			Proposed       & $\mathbf{1.000\pm0.000}$	&$\mathbf{1.000\pm0.000}$	&$\mathbf{1.000\pm0.000}$ &$\mathbf{1.000\pm0.000}$	&$\mathbf{1.000\pm0.000}$	&$\mathbf{1.000\pm0.000}$	&$\mathbf{1.000\pm0.000}$\\  
			\hline\hline
		\end{tabular}
		\label{tab:orl}
	\end{table*}
\begin{table*}
	\caption{Clustering results on \textbf{UCI-digit}. We set $\omega_1=0.4$, $\alpha=4$, and $\lambda=40$ in the proposed method.}
	\centering
	\begin{tabular}{cccccccc}
		\hline\hline
		Method  & ACC & NMI & ARI & F1-score & Precision & Recall & Purity\\
		\hline
		
		L-MSC  \cite{8502831}
		& $0.899\pm0.000$	
		& $0.819\pm0.000$
		& $0.795\pm0.000$	
		& $0.816\pm0.000$	
		& $0.812\pm0.000$	
		& $0.819\pm0.000$	
		& $0.899\pm0.000$
		   \\
		DiMSC  \cite{cao2015diversity}     
		& $0.867\pm0.001$
		& $0.782\pm0.002$
		& $0.747\pm0.002$	
		& $0.772\pm0.002$	
		& $0.769\pm0.002$	
		& $0.775\pm0.002$	
		& $0.867\pm0.001$
		   \\
		
		MCLES \cite{chen2020multi}
		& $0.941\pm	0.004	$
		& $0.891\pm	0.008	$
		& $0.877\pm	0.009	$
		& $0.889\pm	0.008	$
		& $0.885\pm	0.008	$
		& $0.894\pm	0.007	$
		& $0.941\pm	0.004$\\
		AWP \cite{nie2018multiview}
		& $0.871\pm	0.000$	
		& $0.899\pm	0.000$	
		& $0.835\pm	0.000$	
		& $0.853\pm	0.000$	
		& $0.783\pm	0.000$	
		& $0.937\pm	0.000	$
		& $0.872\pm	0.000$\\
		\cdashline{1-8}
		 LT-MSC   \cite{zhang2015low}   
		& $0.792\pm0.009$
		& $0.762\pm0.009$
		& $0.707\pm0.014$
		& $0.737\pm0.013$
		& $0.724\pm0.012$
		& $0.749\pm0.013$
		& $0.809\pm0.009$
		\\
		Ut-SVD-MSC   \cite{IJCVxie2018unifying}
		& $0.804\pm0.001$
		& $0.781\pm0.001	$
		& $0.727\pm0.001	$
		& $0.755\pm0.001	$
		& $0.741\pm0.001	$
		& $0.770\pm0.001	$
		& $0.821\pm0.001$
		  \\
		t-SVD-MSC  \cite{IJCVxie2018unifying}     
		& ${0.966\pm0.001}$
		&	$0.934\pm0.001$
		&	$0.928\pm0.001$
		&	$0.935\pm0.001$
		&	$0.933\pm0.001$
		&	$0.936\pm0.001$
		&	$\mathit{0.966\pm0.001}$
		   \\
		ETLMC \cite{TIPwu2019essential} 
		& $0.941\pm0.023$	
		& $\underline{0.970\pm0.013}	$
		& ${0.933\pm0.029}	$
		& ${0.936\pm0.027}	$
		& ${0.935\pm0.031}	$
		& ${0.938\pm0.024}	$
		& $0.942\pm0.019$
		 \\
		SCMV-3DT \cite{8421595}
		& $0.919\pm0.001	$
		& $0.850\pm	0.001$	
		& $0.833\pm	0.001$	
		& $0.849\pm	0.001$	
		& $0.847\pm	0.001$	
		& $0.852\pm	0.001$	
		& $0.919\pm	0.001$\\
		\cdashline{1-8}
		SNN \cite{Tensor-Low-Rank-2013-TPAMI}
		& $0.966\pm	0.001	$
		& $0.934\pm	0.001$	
		& $0.928\pm	0.001$	
		& $0.935\pm	0.001$	
		& $0.933\pm	0.001$	
		& $0.936\pm	0.001$	
		& $0.966\pm	0.001$\\
		TRPCA \cite{8606166}
		& $\underline{0.977\pm	0.000}	$
		& $0.948\pm	0.000$	
		& $\underline{0.949\pm	0.000}$	
		& $\underline{0.954\pm	0.000}$	
		& $\underline{0.954\pm	0.000}$	
		& $\underline{0.955\pm	0.000}$	
		& $\underline{0.977\pm	0.000}$\\
		\cdashline{1-8}
		Proposed       
		& $\mathbf{0.998\pm0.000}$ 
		&	$\mathbf{0.993\pm0.000}$
		&	$\mathbf{0.994\pm0.000}$
		&	$\mathbf{0.995\pm0.000}$
		&	$\mathbf{0.995\pm0.000}$
		&	$\mathbf{0.995\pm0.000}$
		&	$\mathbf{0.998\pm0.000}$
		  \\
		\hline\hline
	\end{tabular}
	\label{tab:UCI-digital}
\end{table*}

Five  commonly-used multi-view clustering image datasets were employed to evaluate the proposed model, including: 

\begin{itemize}
    \item \textbf{Coil20} is an image dataset with 20 objects, where each object has 72 samples. 
    \item \textbf{Yale} consists of 165 face images from 15 individuals.
    \item \textbf{YaleB} is a face image dataset with 38 individuals, where each individual has approximate 64 images. As done in \cite{IJCVxie2018unifying}, we used the first 10 classes. Due to the large variation of luminance, clustering on YaleB is quite challenging.
    \item \textbf{ORL} is a face image datasets with 40 individuals, where each individual consists of 10 images. 
    \item \textbf{UCI-digit} contains 2000 handwritten digits corresponding to 10 classes. 
\end{itemize}
Fig. \ref{fig:samples} shows some examples of the adopted datasets. To construct the multi-view data, for all the face and object datasets, three types of features were extracted, i.e., Gabor \cite{lyons1998coding}, LBP \cite{ahonen2006face}, and intensity.  
Specifically, the LBP features were extracted with the sampling density of size  8 and the uniform LBP histogram is in an (8,1) neighbourhood, and the Gabor features were extracted with one scale, 4 orientations, 39 rows and 39 columns in a 2-D garbor filter.
For UCI-digit, we adopted the same features as \cite{TIPwu2019essential},
i.e., the Fourier coefficients, the morphological features and the pixel averages, to construct 3 views. 
Table 1 summarizes the main information of the employed datasets.

\begin{figure*}[!t]
	\begin{minipage}[b]{0.195\linewidth}
		\centering
		\centerline{\epsfig{figure=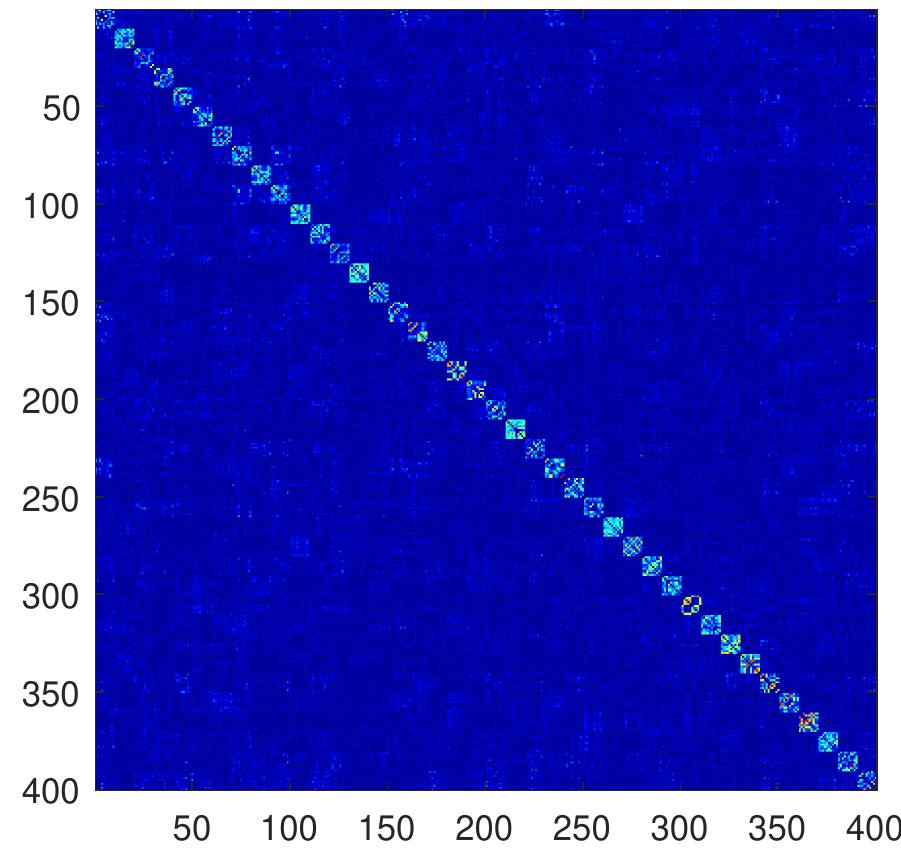,width=3.5cm}}
		\centerline{ORL: LT-MSC}\medskip
	\end{minipage}
	\begin{minipage}[b]{0.195\linewidth}
		\centering
		\centerline{\epsfig{figure=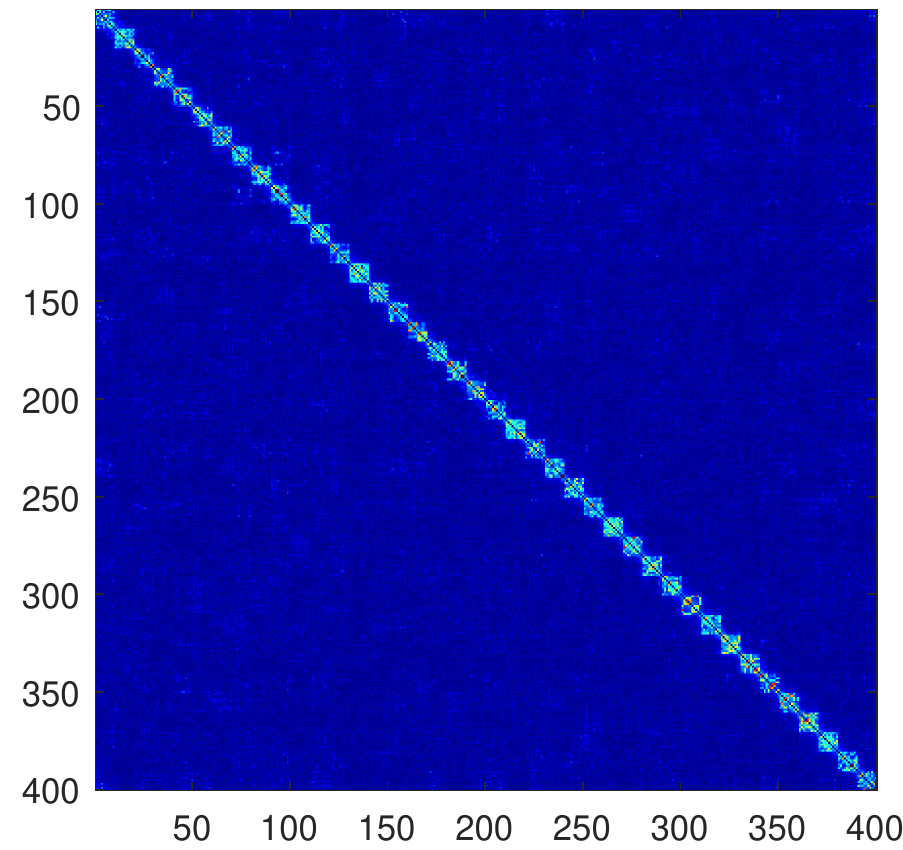,width=3.5cm}}
		\centerline{ORL: t-SVD-MSC}\medskip
	\end{minipage}
	\begin{minipage}[b]{0.195\linewidth}
		\centering
		\centerline{\epsfig{figure=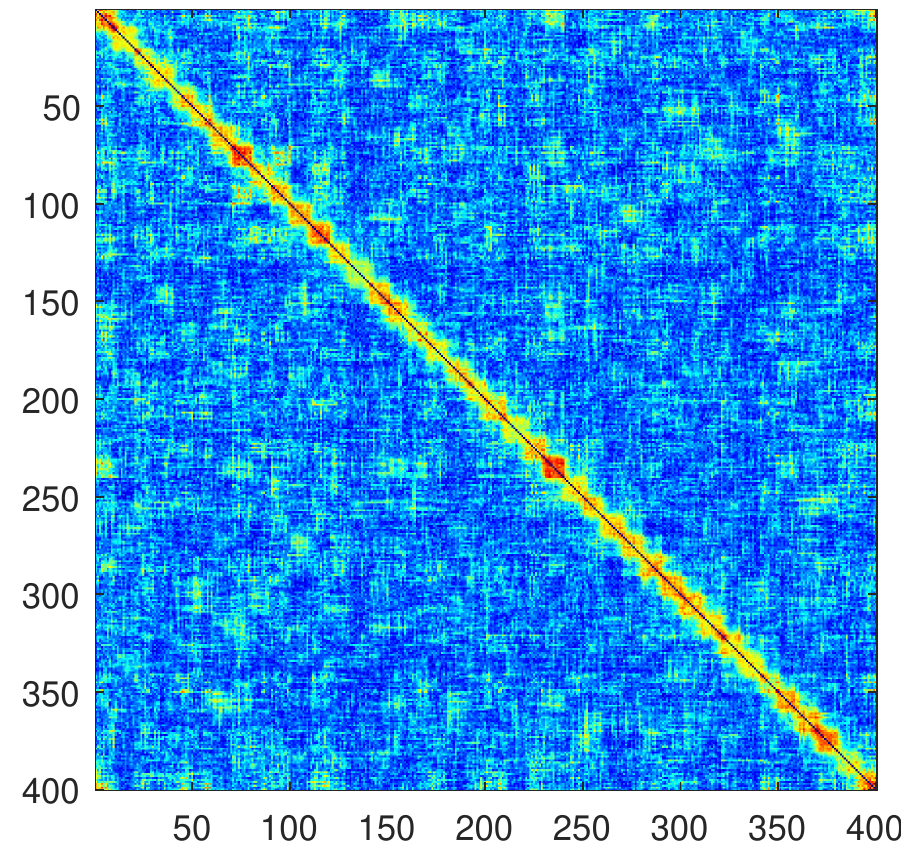,width=3.5cm}}
		\centerline{ORL: SNN}\medskip
	\end{minipage}
	\begin{minipage}[b]{0.195\linewidth}
		\centering
		\centerline{\epsfig{figure=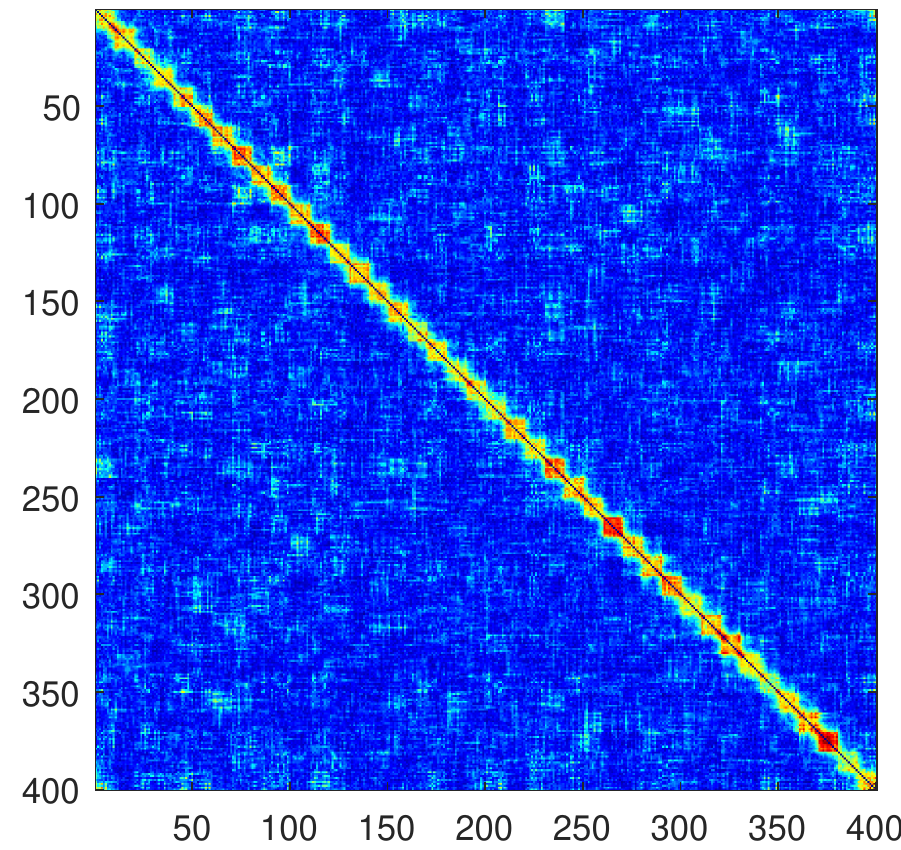,width=3.5cm}}
		\centerline{ORL: TRPCA}\medskip
	\end{minipage}
	\begin{minipage}[b]{0.195\linewidth}
		\centering
		\centerline{\epsfig{figure=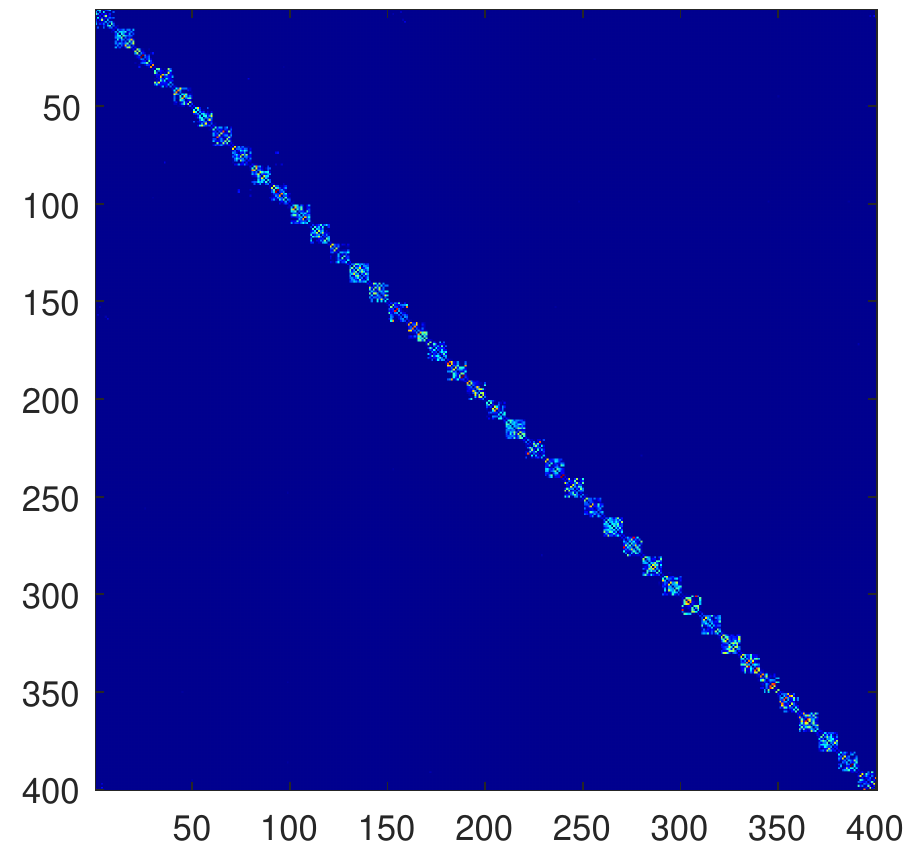,width=3.5cm}}
		\centerline{ORL: Proposed}\medskip
	\end{minipage}\\
	\begin{minipage}[b]{0.195\linewidth}
		\centering
		\centerline{\epsfig{figure=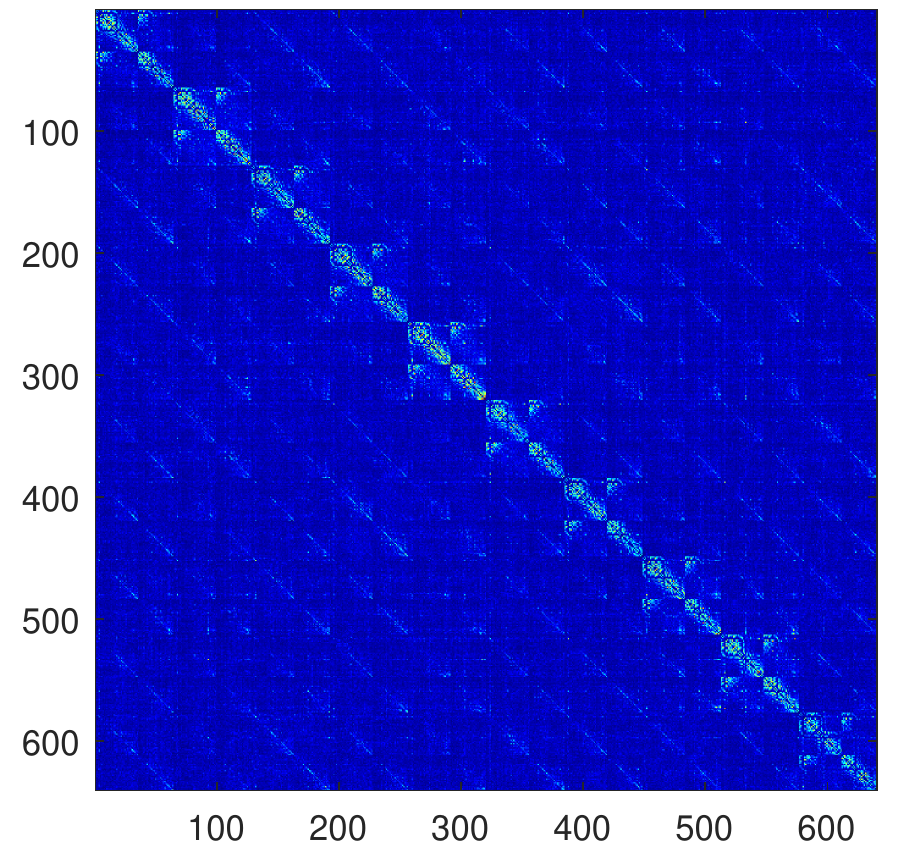,width=3.5cm}}
		\centerline{YaleB: LT-MSC}\medskip
	\end{minipage}
	\begin{minipage}[b]{0.195\linewidth}
		\centering
		\centerline{\epsfig{figure=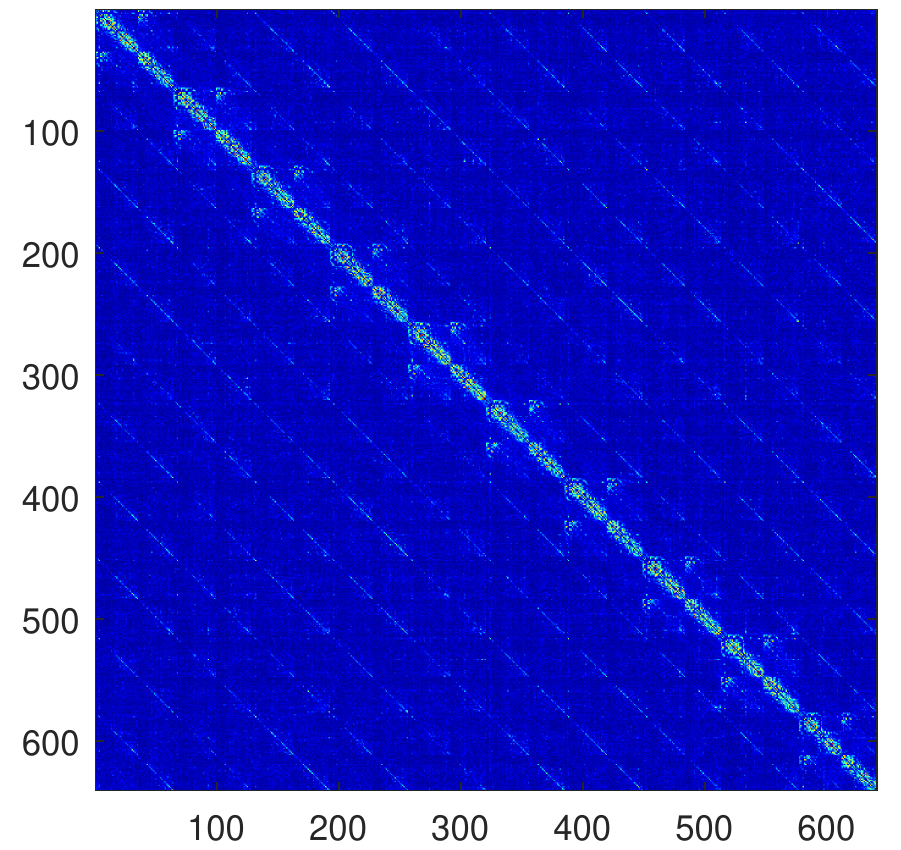,width=3.5cm}}
		\centerline{YaleB: t-SVD-MSC}\medskip
	\end{minipage}
	\begin{minipage}[b]{0.195\linewidth}
		\centering
		\centerline{\epsfig{figure=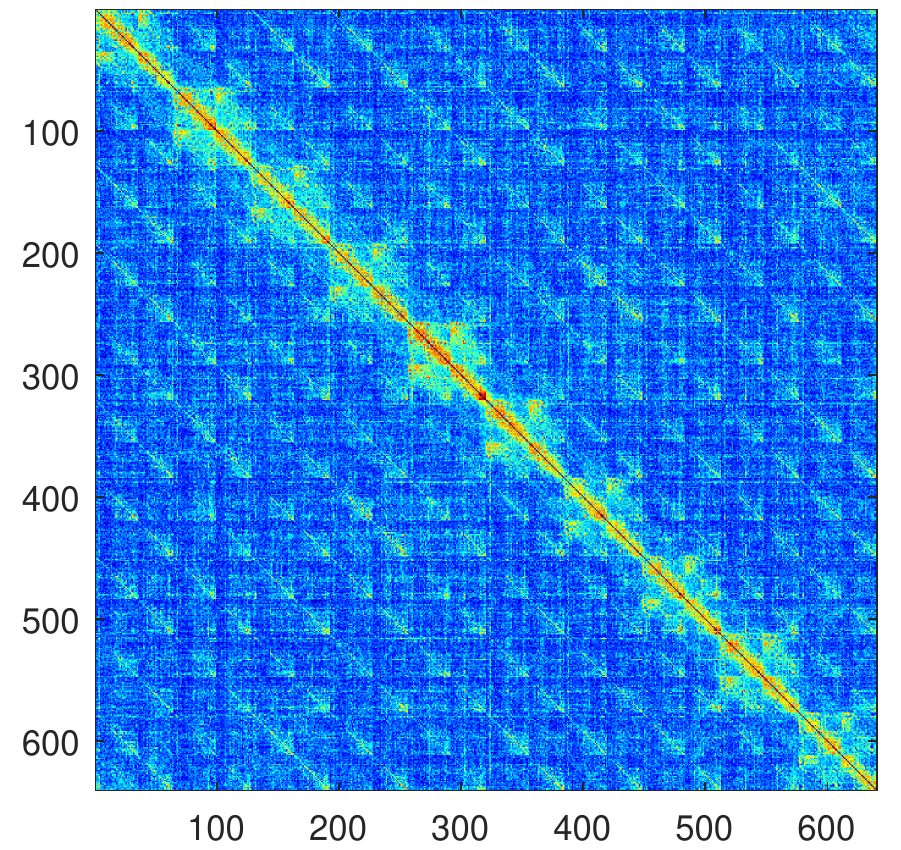,width=3.5cm}}
		\centerline{YaleB: SNN}\medskip
	\end{minipage}
	\begin{minipage}[b]{0.195\linewidth}
		\centering
		\centerline{\epsfig{figure=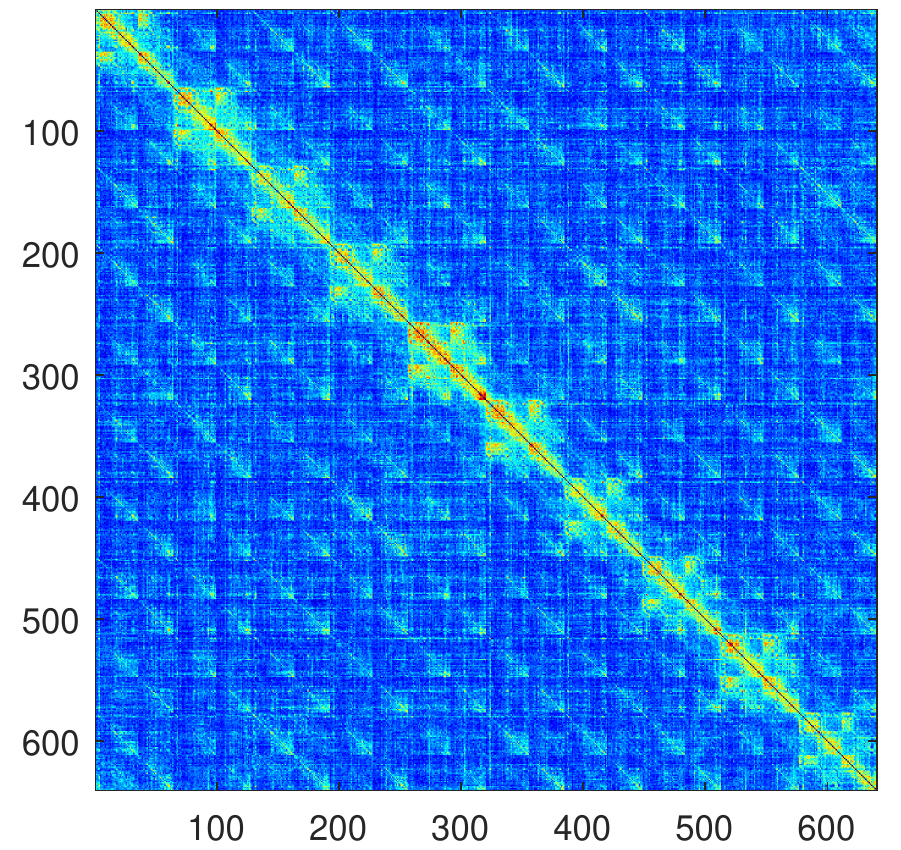,width=3.5cm}}
		\centerline{YaleB: TRPCA}\medskip
	\end{minipage}
	\begin{minipage}[b]{0.195\linewidth}
		\centering
		\centerline{\epsfig{figure=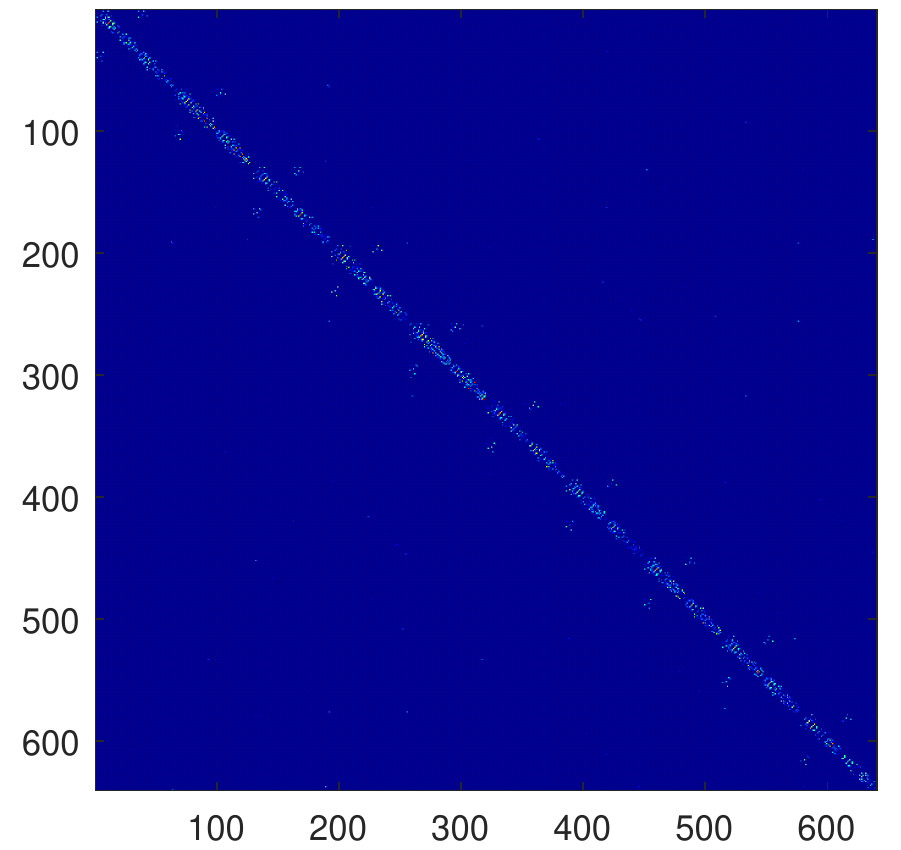,width=3.5cm}}
		\centerline{YaleB: Proposed}\medskip
	\end{minipage}
	\caption{Visual comparisons  of the similarity matrices learned by different methods.  
	It is clear that the connections of our method are distributed along the diagonal,  
	and our method generates much fewer wrong connections than the compared methods.}
	\label{fig:visual}
\end{figure*}

\begin{figure*}[!t]
	\begin{minipage}[b]{0.162\linewidth}
		\centering
		\centerline{\epsfig{figure=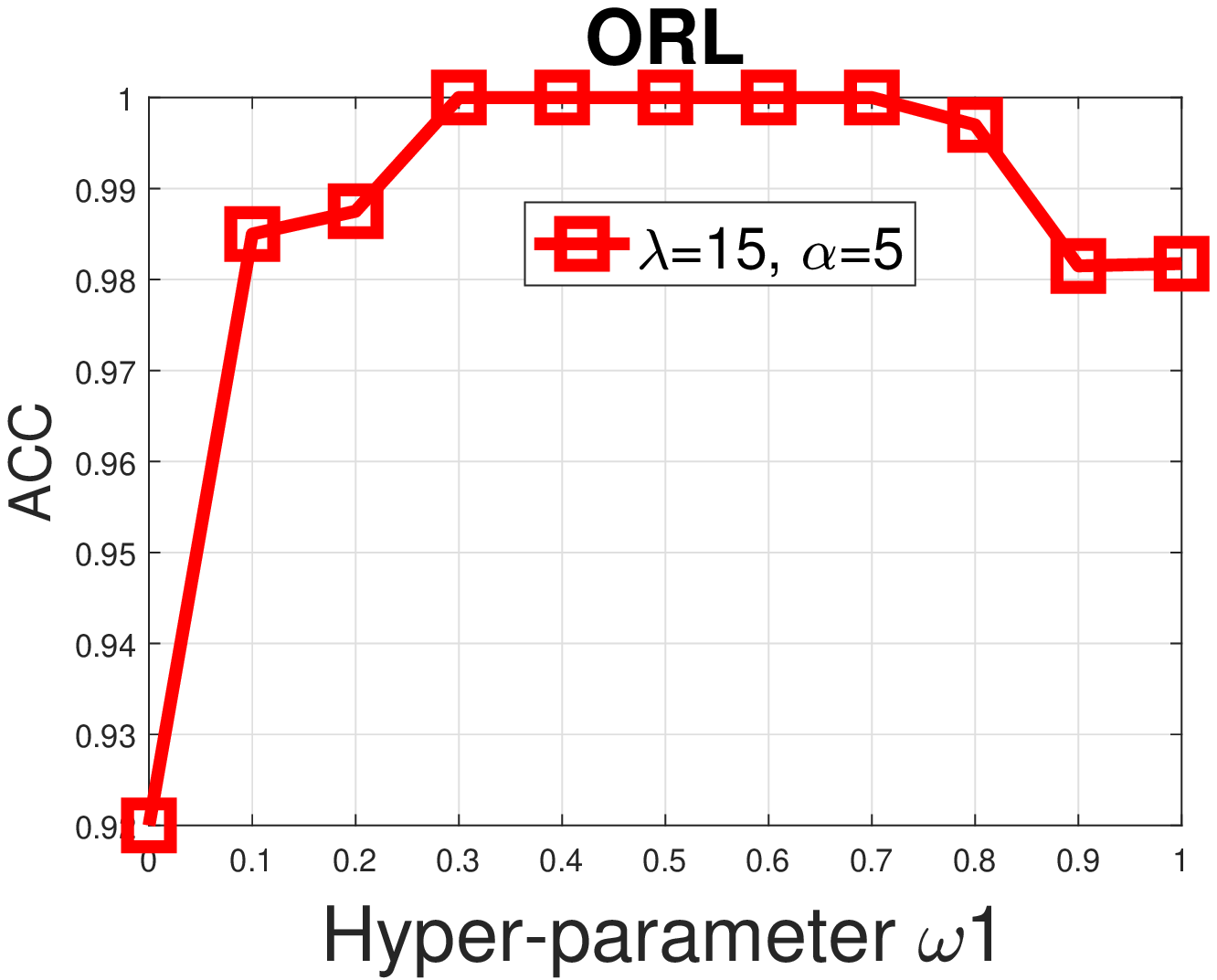,width=3.2cm}}
	\end{minipage}
	\begin{minipage}[b]{0.162\linewidth}
		\centering
		\centerline{\epsfig{figure=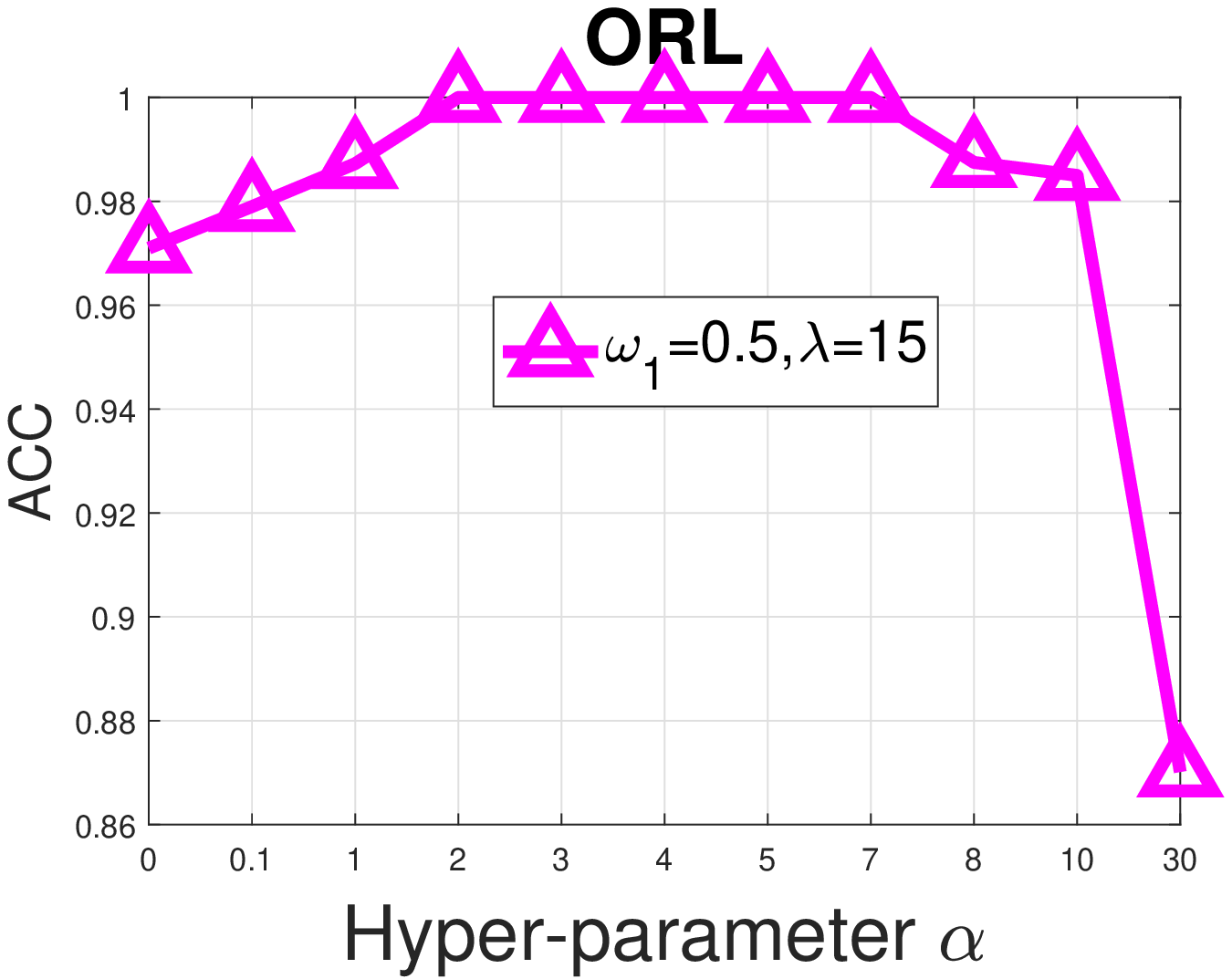,width=3.2cm}}
	\end{minipage}
	\begin{minipage}[b]{0.162\linewidth}
		\centering
		\centerline{\epsfig{figure=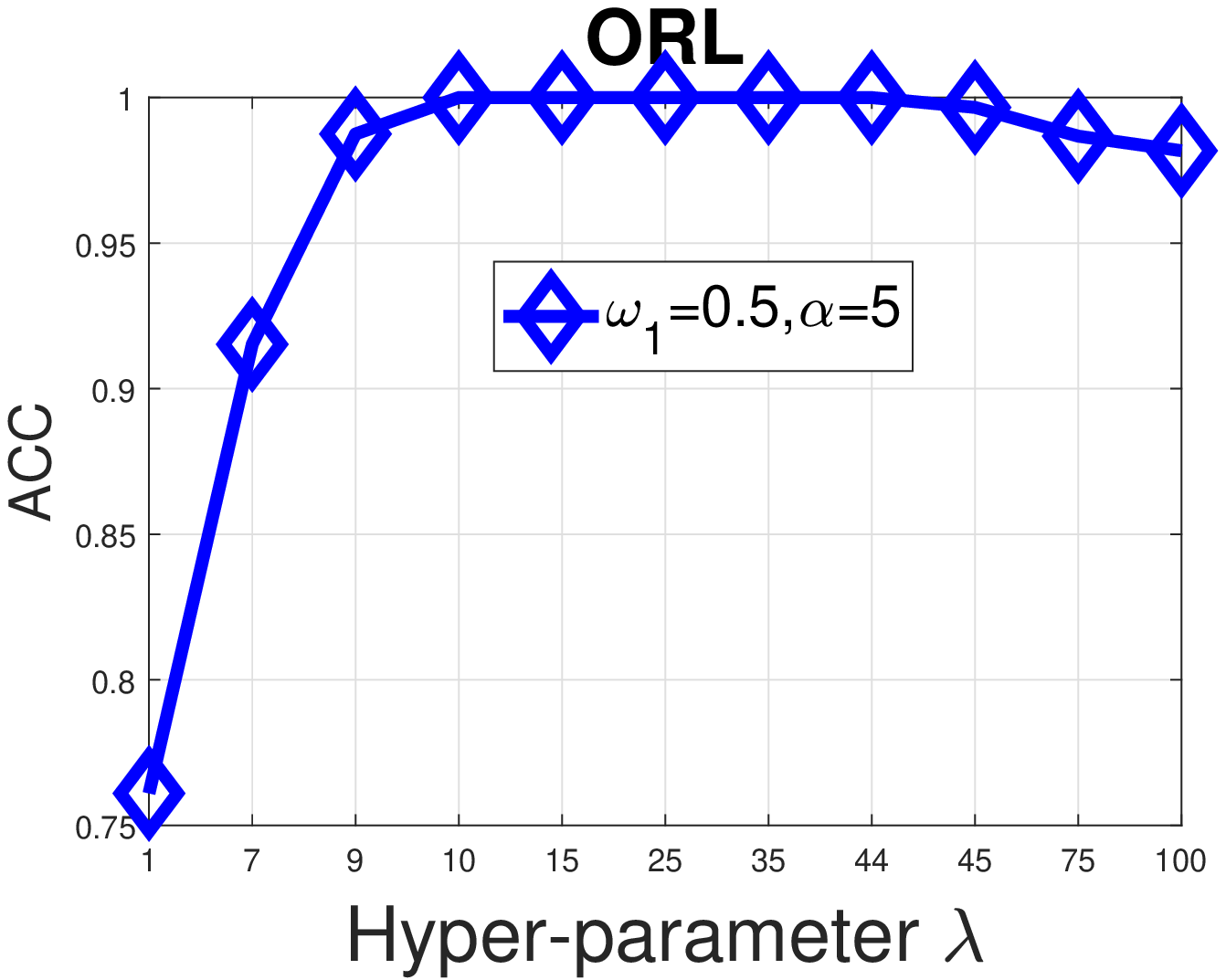,width=3.2cm}}
	\end{minipage}
	\begin{minipage}[b]{0.162\linewidth}
		\centering
		\centerline{\epsfig{figure=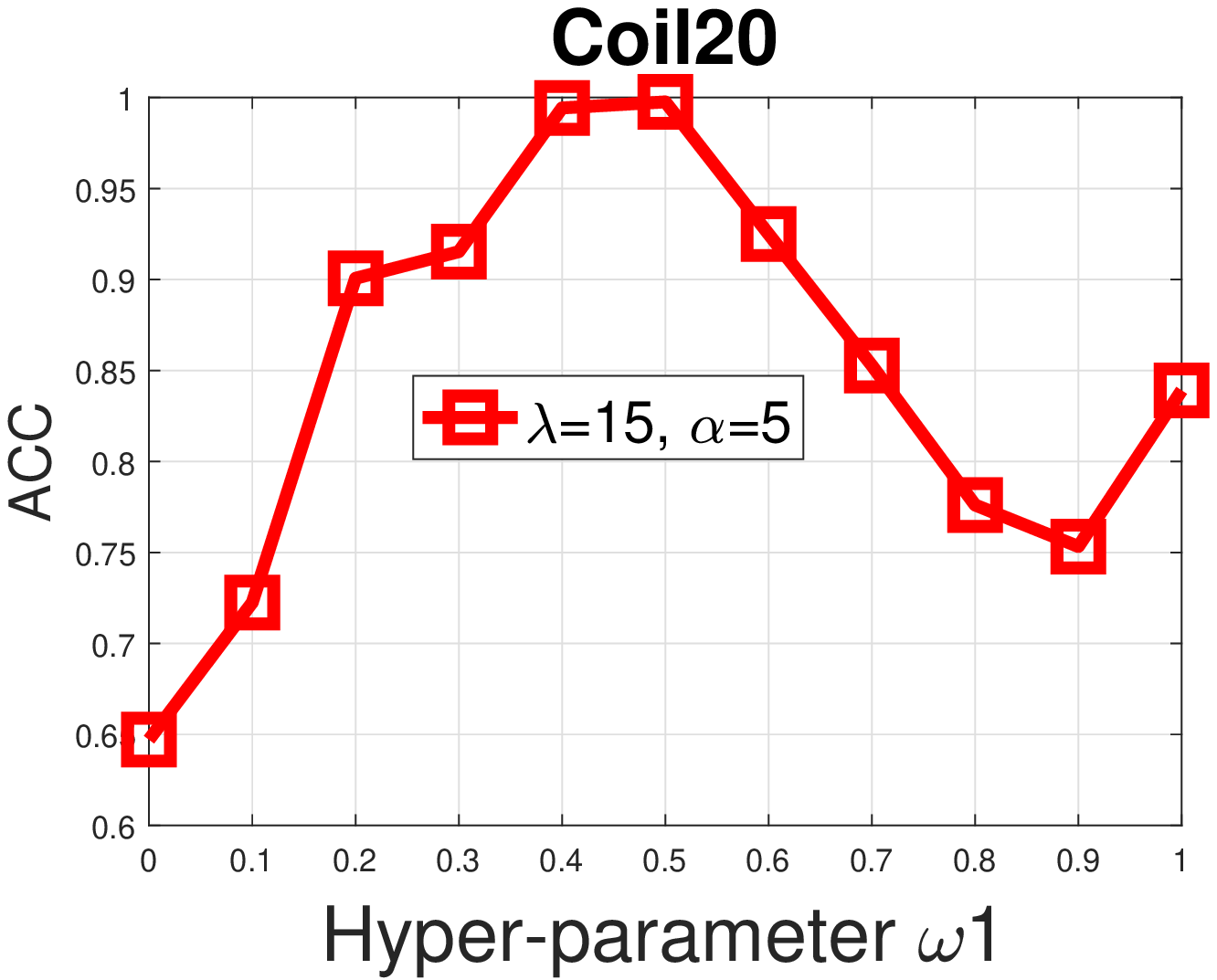,width=3.2cm}}
	\end{minipage}
	\begin{minipage}[b]{0.162\linewidth}
		\centering
		\centerline{\epsfig{figure=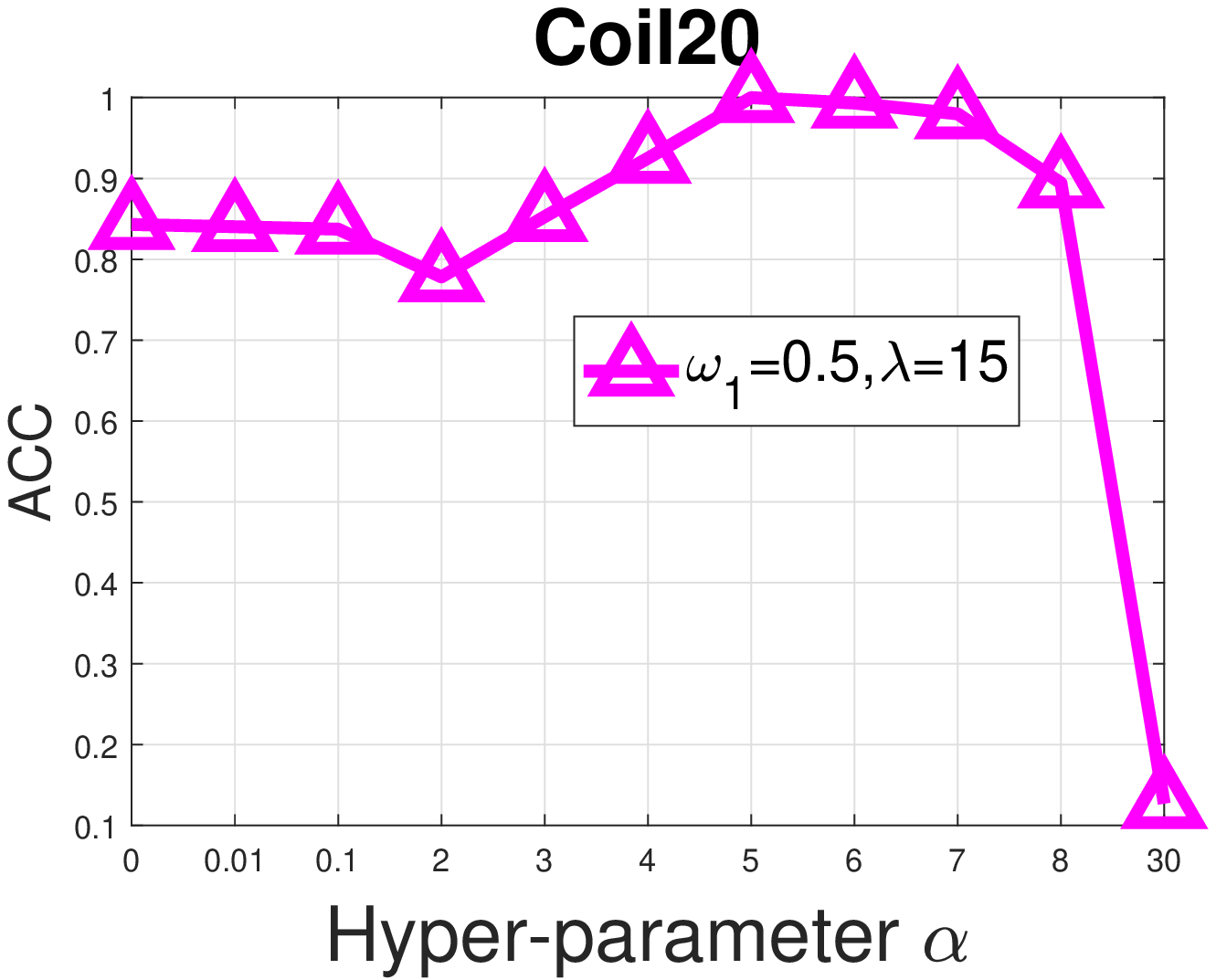,width=3.2cm}}
	\end{minipage}
	\begin{minipage}[b]{0.162\linewidth}
		\centering
		\centerline{\epsfig{figure=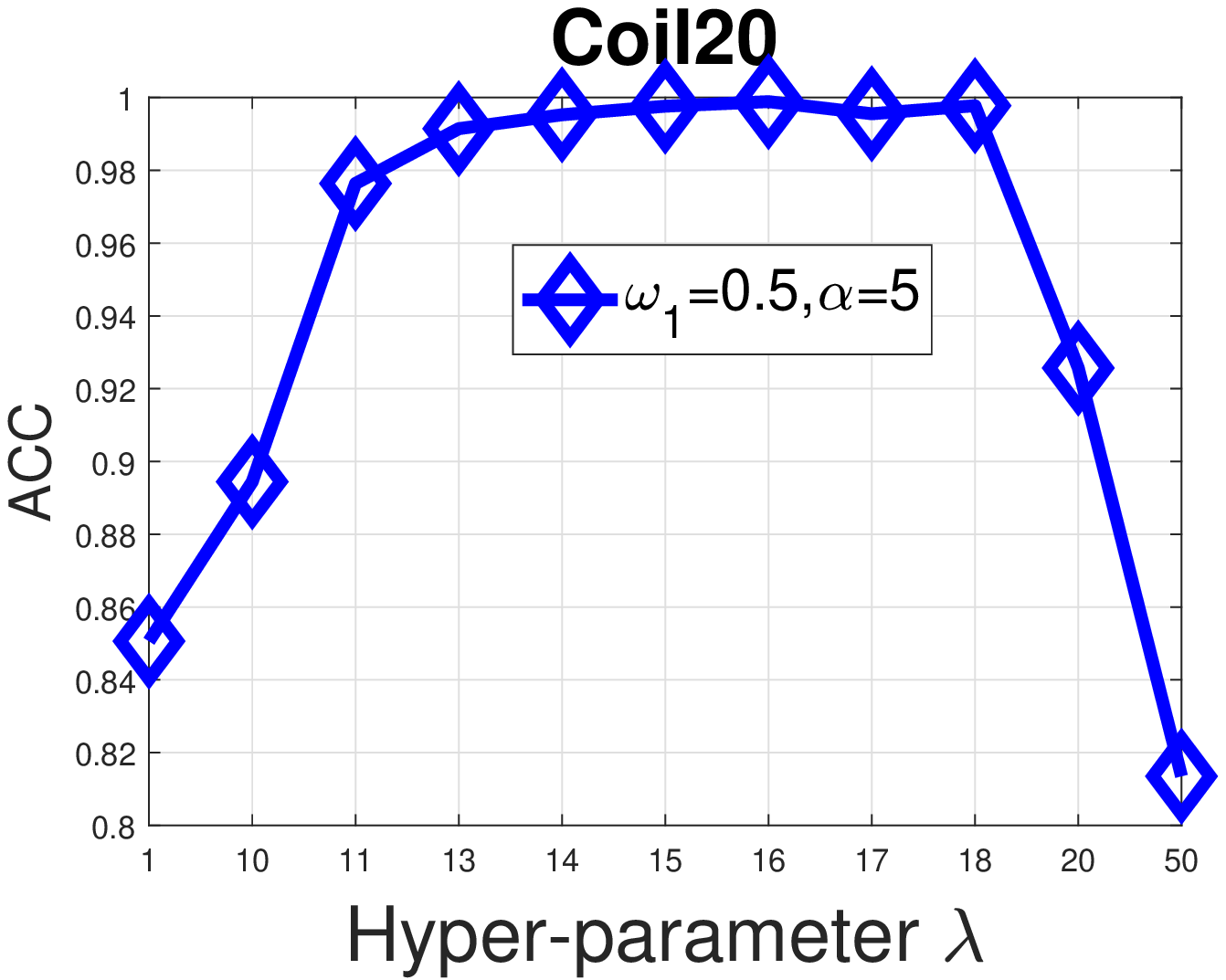,width=3.2cm}}
	\end{minipage}
	\caption{Illustration of the effect of three hyper-parameters of our model (i.e., $\omega_1,\lambda$ and $\alpha$) on the clustering performance. It can be observed that the proposed method always achieves the highest performance in wide ranges of $\omega_1,\lambda$ and $\alpha$, demonstrating its robustness.}
	\label{fig:para}
\end{figure*}

We compared the proposed method with four  state-of-the-art non-tensor based MVSC methods, i.e., 
\begin{itemize}
    \item[1.] \textbf{L-MSC} [2020, TPAMI] \cite{8502831} learns a shared latent representation from all the views and  exploits the complementary information from different views simultaneously;
    \item[2.] \textbf{DiMSC} [2015, CVPR] \cite{cao2015diversity}  learns  highly diverse representations for each view to emphasize the complementary information;
    \item[3.] \textbf{MCLES} [2020, AAAI] \cite{chen2020multi} is a multi-view clustering method in latent space. Besides, it could directly produce the clustering result without spectral clustering;  and 
    \item[4.] \textbf{AWP} [2018, SIGKDD] \cite{nie2018multiview} can automatically determine the importance of each view and achieve MVSC by spectral rotation. 
\end{itemize}
We also compared with five state-of-the-art tensor based MVSC methods:
\begin{itemize}
    \item[5.] \textbf{LT-MSC} [2015, ICCV] \cite{zhang2015low} is low-rank tensor constrained self-representative model for MVSC, which uses the  TLRN from  \cite{Tensor-Low-Rank-2013-TPAMI}; 
    \item[6.] \textbf{Ut-SVD-MSC} [2018, IJCV] \cite{IJCVxie2018unifying} is an extension of LT-MSC with a different TLRN based on t-SVD \cite{SVD-tensor}; 
    \item[7.] \textbf{t-SVD-MSC} [2018, IJCV] \cite{IJCVxie2018unifying} promotes the performance of Ut-SVD-MSC by rotating the original tensor to ensure the consensus among multiple views; 
    \item[8.] \textbf{ETLMSC} [2019, TIP] \cite{TIPwu2019essential} is an essential tensor learning method to explore the high-order correlations for multi-view representations; and 
    \item[9.] \textbf{SCMV-3DT} [2019, TNNLS] \cite{8421595} uses the tensor-to-tensor product to learn the similarity matrix for MVSC. 
\end{itemize}
Moreover, to illustrate the advantage of the proposed TLRN over the existing TLRNs, we also compared with two popular TLRNs: 
\begin{itemize}
    \item[10.] \textbf{SNN} [2013, TPAMI] \cite{Tensor-Low-Rank-2013-TPAMI} denotes the  sum of nuclear norms, which is a relaxation of tensor Tucker rank; and 
    \item[11.] \textbf{TRPCA} [2020, TPAMI] \cite{8606166} represents the recently proposed TLRNs induced by t-SVD. 
\end{itemize}
The codes of all the methods under comparison  are provided by the authors. 
For all the methods, we used the SC method in \cite{ng2002spectral} to perform clustering. 
For the compared MVSC methods, we exhaustively turned the hyper-parameters according to the suggested ranges  by the original papers, and reported the best average results and the standard deviations (std) on 20 trails. 
For the proposed method and two compared TLRNs, the initial similarity matrices of different views were initialized as those  from \cite{IJCVxie2018unifying}. Note that  other similarity matrix construction methods can also be adopted to initialize  the proposed model.

Seven  metrics were adopted to evaluate the clustering results, i.e., clustering accuracy (ACC), 
normalized mutual information (NMI), adjusted rank index (ARI), F1-score, Precision, Recall and Purity.
ARI lies  in the range of $[-1,1]$, and all the remaining  metrics lie  in the range of $[0,1]$.
For all metrics, larger values indicate better clustering performance, and when perfect clustering is achieved, the metrics will reach $1$.

\begin{table*}
	\caption{Ablation study towards the effectiveness of the regularization terms contained in our method. When the parameter is equal to zero, the corresponding term is removed from our model.}
	\centering
	\scalebox{1}{
		\begin{tabular}{cccccccc}
			\hline\hline
			\textbf{Coil20}	&	ACC		&		NMI		&		ARI		&		F1-score		&		Precision		&		Recall		&		Purity		\\ \hline	
			
			$\alpha=0$	&$	0.848 	\pm	0.018 	$&$	0.928 	\pm	0.005 	$&$	0.815 	\pm	0.019 	$&$	0.825 	\pm	0.018 	$&$	0.780 	\pm	0.031 	$&$	0.875 	\pm	0.008 	$&$	0.885 	\pm	0.009 	$\\
			$\omega_1=0$	&$	0.647 	\pm	0.010 	$&$	0.825 	\pm	0.005 	$&$	0.409 	\pm	0.006 	$&$	0.451 	\pm	0.005 	$&$	0.310 	\pm	0.004 	$&$	0.828 	\pm	0.012 	$&$	0.721 	\pm	0.007 	$\\
			$\omega_2=0$	&$	0.842 	\pm	0.013 	$&$	0.928 	\pm	0.004 	$&$	0.810 	\pm	0.018 	$&$	0.820 	\pm	0.017 	$&$	0.771 	\pm	0.028 	$&$	0.876 	\pm	0.006 	$&$	0.884 	\pm	0.006 	$\\
			Proposed	&$	\mathbf{1.000 	\pm	0.000} 	$&$	\mathbf{1.000 	\pm	0.000} 	$&$	\mathbf{1.000 	\pm	0.000} 	$&$	\mathbf{1.000 	\pm	0.000} 	$&$	\mathbf{1.000 	\pm	0.000} 	$&$	\mathbf{1.000 	\pm	0.000} 	$&$	\mathbf{1.000 	\pm	0.000} 	$\\ \hline\hline
			
			\textbf{Yale}	&	ACC		&		NMI		&		ARI		&		F1-score		&		Precision		&		Recall		&		Purity		\\ \hline	
			$\alpha=0$	&$	0.849 	\pm	0.036 	$&$	0.883 	\pm	0.027 	$&$	0.743 	\pm	0.062 	$&$	0.759 	\pm	0.058 	$&$	0.725 	\pm	0.067 	$&$	0.796 	\pm	0.047 	$&$	0.851 	\pm	0.035 	$\\
			$\omega_1=0$	&$	0.686 	\pm	0.011 	$&$	0.747 	\pm	0.006 	$&$	0.520 	\pm	0.016 	$&$	0.552 	\pm	0.014 	$&$	0.506 	\pm	0.020 	$&$	0.607 	\pm	0.014 	$&$	0.703 	\pm	0.009 	$\\
			$\omega_2=0$	&$	0.849 	\pm	0.037 	$&$	0.882 	\pm	0.029 	$&$	0.736 	\pm	0.068 	$&$	0.753 	\pm	0.064 	$&$	0.718 	\pm	0.075 	$&$	0.794 	\pm	0.050 	$&$	0.849 	\pm	0.037 	$\\
			Proposed	&$	\mathbf{0.988 	\pm	0.001} 	$&$	\mathbf{0.988 	\pm	0.001} 	$&$	\mathbf{0.975 	\pm	0.003} 	$&$	\mathbf{0.976 	\pm	0.003} 	$&$	\mathbf{0.974 	\pm	0.003} 	$&$	\mathbf{0.979 	\pm	0.002} 	$&$	\mathbf{0.988 	\pm	0.001} 	$\\ \hline\hline
			\textbf{ORL}	&	ACC		&		NMI		&		ARI		&		F1-score		&		Precision		&		Recall		&		Purity		\\ \hline	
			$\alpha=0$	&$	0.979 	\pm	0.016 	$&$	0.996 	\pm	0.003 	$&$	0.979 	\pm	0.016 	$&$	0.979 	\pm	0.016 	$&$	0.966 	\pm	0.026 	$&$	0.993 	\pm	0.006 	$&$	0.984 	\pm	0.012 	$\\
			$\omega_1=0$	&$	0.918 	\pm	0.000 	$&$	0.977 	\pm	0.001 	$&$	0.897 	\pm	0.009 	$&$	0.900 	\pm	0.009 	$&$	0.852 	\pm	0.016 	$&$	0.954 	\pm	0.000 	$&$	0.940 	\pm	0.000 	$\\
			$\omega_2=0$	&$	0.980 	\pm	0.017 	$&$	0.996 	\pm	0.003 	$&$	0.980 	\pm	0.017 	$&$	0.980 	\pm	0.017 	$&$	0.968 	\pm	0.027 	$&$	0.993 	\pm	0.006 	$&$	0.985 	\pm	0.013 	$\\
			Proposed	&$	\mathbf{1.000 	\pm	0.000} 	$&$\mathbf{	1.000 	\pm	0.000} 	$&$	\mathbf{1.000 	\pm	0.000} 	$&$	\mathbf{1.000 	\pm	0.000} 	$&$	\mathbf{1.000 	\pm	0.000} 	$&$	\mathbf{1.000 	\pm	0.000} 	$&$	\mathbf{1.000 	\pm	0.000} 	$\\ \hline\hline
			\textbf{UCI-digital}	&	ACC		&		NMI		&		ARI		&		F1-score		&		Precision		&		Recall		&		Purity		\\ \hline	
			$\alpha=0$	&$	0.996 	\pm	0.000 	$&$	0.988 	\pm	0.000 	$&$	0.990 	\pm	0.000 	$&$	0.991 	\pm	0.000 	$&$	0.991 	\pm	0.000 	$&$	0.991 	\pm	0.000 	$&$	0.996 	\pm	0.000 	$\\
			$\omega_1=0$	&$	0.987 	\pm	0.000 	$&$	0.973 	\pm	0.000 	$&$	0.970 	\pm	0.000 	$&$	0.973 	\pm	0.000 	$&$	0.973 	\pm	0.000 	$&$	0.974 	\pm	0.000 	$&$	0.987 	\pm	0.000 	$\\
			$\omega_2=0$	&$	0.996 	\pm	0.000 	$&$	0.988 	\pm	0.000 	$&$	0.990 	\pm	0.000 	$&$	0.991 	\pm	0.000 	$&$	0.991 	\pm	0.000 	$&$	0.991 	\pm	0.000 	$&$	0.996 	\pm	0.000 	$\\
			Proposed	&$	\mathbf{0.998 	\pm	0.000} 	$&$	\mathbf{0.993 	\pm	0.000} 	$&$	\mathbf{0.994 	\pm	0.000} 	$&$	\mathbf{0.995 	\pm	0.000} 	$&$	\mathbf{0.995 	\pm	0.000} 	$&$	\mathbf{0.995 	\pm	0.000} 	$&$	\mathbf{0.998 	\pm	0.000} 	$\\ \hline\hline
			\textbf{YaleB}	&	ACC		&		NMI		&		ARI		&		F1-score		&		Precision		&		Recall		&		Purity		\\ \hline	
			$\alpha=0$	&$	0.605 	\pm	0.003 	$&$	0.619 	\pm	0.004 	$&$	0.399 	\pm	0.006 	$&$	0.463 	\pm	0.005 	$&$	0.427 	\pm	0.005 	$&$	0.505 	\pm	0.004 	$&$	0.607 	\pm	0.003 	$\\
			$\omega_1=0$	&$	0.138 	\pm	0.003 	$&$	0.090 	\pm	0.003 	$&$	0.001 	\pm	0.000 	$&$	0.178 	\pm	0.000 	$&$	0.099 	\pm	0.000 	$&$	0.869 	\pm	0.003 	$&$	0.153 	\pm	0.003 	$\\
			$\omega_2=0$	&$	0.609 	\pm	0.003 	$&$	0.624 	\pm	0.004 	$&$	0.405 	\pm	0.005 	$&$	0.468 	\pm	0.005 	$&$	0.433 	\pm	0.005 	$&$	0.510 	\pm	0.004 	$&$	0.610 	\pm	0.003 	$\\
			Proposed	&$\mathbf{0.954\pm0.002}$
			& $\mathbf{0.908\pm0.001}$
			& $\mathbf{0.900\pm0.003}$	
			& $\mathbf{0.910\pm0.002}$
			& $\mathbf{0.909\pm0.003}$	
			& $\mathbf{0.912\pm0.003}$	
			& $\mathbf{0.954\pm0.003}$\\
			\hline\hline
	\end{tabular}}
	\label{tab:abS}
\end{table*}

\begin{figure}
    \centering
	\centerline{\epsfig{figure=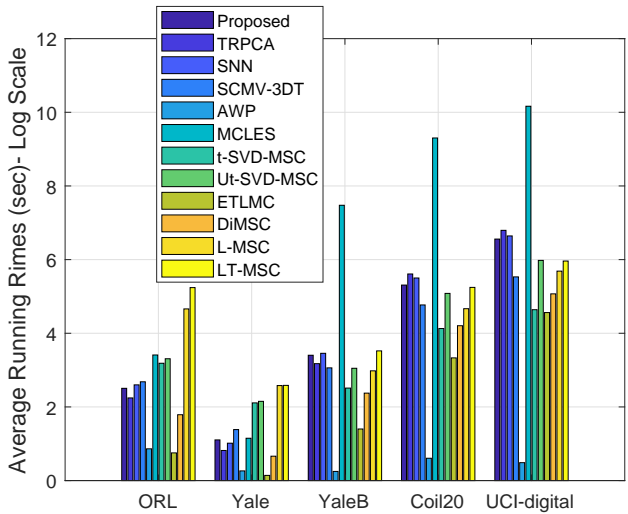,width=9.6cm}}
    \caption{Running time comparison of differnet methods on five datasets.}
    \label{fig:Running_time}
\end{figure}
\subsection{Comparison of Clustering Performance}

Tables \ref{tab:coil20}-\ref{tab:UCI-digital} list clustering results  of all the methods, where the most impressive observation  is that our method produces  the perfect clustering results on Coil20 and ORL, and almost  perfect clustering results on Yale and UCI-digit. Specifically, on Yale, the ACC value  of our model is $0.988$, which means there are only two samples being wrongly partitioned by the proposed method on average. On UCI-digit, the ACC value is $0.998$, indicating that  only 4 samples are assigned to the incorrect groups among 2000 samples. On ORL and Coil20, all the samples can be clustered correctly. Clustering on YaleB is a quite challenging task due to the large variation of luminance. Fortunately, our method also achieves the best and surprising  performance, i.e., all the clustering metrics exceed $0.90$. 
On the contrary, all the metrics for the other methods are lower than $0.71$.  This means a dramatic improvement of the proposed model over the compared methods, e.g.,  our method improves the ARI value  more than $65\%$ when compared with the second best method. 
Moreover, the improvements of the proposed method on the other datasets are also significant.
Different methods have different assumptions for input data, and may favor different datasets. For example, t-SVD-MSC achieves high values of various metrics  on Yale and ORL, but relatively low values of those on Coil20. ETLMC performs good on UCI-digit, but bad on YaleB. 
\textit{Remarkably, our method consistently produces the best  performance on all the datasets, validating  the robustness of our methods to different datasets}.   
In addition, our method also significantly outperforms SNN and TRPCA. The reason is that they are designed for general purposes without considering the special characteristics of MVSC, while the proposed TLRN is tailored to MVSC, leading to the superior clustering performance.



%
%
%
%
%
%
%
%
%
%
%




\begin{table*}
	\caption{Clustering Performance of Our Model with Different Initial Affinity Matrices}
	\centering
	\scalebox{0.95}{
		\begin{tabular}{cccccccc}
			\hline\hline	
			\textbf{ORL}	&	ACC		&		NMI		&		ARI		&		F1-score		&		Precision		&		Recall		&		Purity		\\ \hline	
			LT-MSC \cite{zhang2015low}	&$	0.808 	\pm	0.022 	$&$	0.915 	\pm	0.010 	$&$	0.747 	\pm	0.029 	$&$	0.753 	\pm	0.028 	$&$	0.708 	\pm	0.034 	$&$	0.805 	\pm	0.023 	$&$	0.838 	\pm	0.015 	$\\
			Proposed (LT-MSC)	&$	\mathbf{0.871 	\pm	0.018} 	$&$	\mathbf{0.950 	\pm	0.005} 	$&$	\mathbf{0.838 	\pm	0.015} 	$&$	\mathbf{0.841 	\pm	0.014} 	$&$	\mathbf{0.801 	\pm	0.020} 	$&$	\mathbf{0.886 	\pm	0.010} 	$&$	\mathbf{0.901 	\pm	0.013} 	$\\ \cdashline{1-8}
			ETLMC  \cite{TIPwu2019essential}	&$0.946\pm0.018$
			& $0.986\pm0.005$
			& $0.942\pm0.020$
			& $0.943\pm0.019$
			& $0.918\pm0.026$
			& $0.970\pm0.014$
			& $0.960\pm0.014$\\
			Proposed (ETLMC)	&$	\mathbf{0.960 	\pm	0.008} 	$&$	\mathbf{0.987 	\pm	0.003} 	$&$	\mathbf{0.952 	\pm	0.009} 	$&$	\mathbf{0.953 	\pm	0.009} 	$&$	\mathbf{0.936 	\pm	0.011} 	$&$	\mathbf{0.972 	\pm	0.008} 	$&$	\mathbf{0.969 	\pm	0.005} 	$\\ 
			\hline\hline
			\textbf{YaleB}	&	ACC		&		NMI		&		ARI		&		F1-score		&		Precision		&		Recall		&		Purity		\\ \hline	
			LT-MSC \cite{zhang2015low}	&$0.617\pm0.002$
			& $0.623\pm0.006$
			& $0.413\pm0.006$
			& $0.491\pm0.005$
			& $0.466\pm0.005$
			& $0.521\pm0.005$
			& $0.620\pm0.002$\\
			Proposed (LT-MSC)	&$	\mathbf{0.831 	\pm	0.002} 	$&$	\mathbf{0.778 	\pm	0.001} 	$&$	\mathbf{0.593 	\pm	0.003} 	$&$	\mathbf{0.638 	\pm	0.002} 	$&$	\mathbf{0.571 	\pm	0.003} 	$&$	\mathbf{0.723 	\pm	0.002} 	$&$	\mathbf{0.831 	\pm	0.002} 	$\\ \cdashline{1-8}
			ETLMC  \cite{TIPwu2019essential}	&$0.325\pm0.011$
			& $0.307\pm0.021$
			& $\mathbf{0.179\pm0.019}$
			& $0.262\pm0.017$
			& $\mathbf{0.257\pm0.017}$
			& $0.267\pm0.017$
			& $0.332\pm0.010$\\
			Proposed (ETLMC)	&$	\mathbf{0.452 	\pm	0.000} 	$&$	\mathbf{0.455 	\pm	0.000} 	$&$	0.177 	\pm	0.000 	$&$	\mathbf{0.278 	\pm	0.000} 	$&$	0.224 	\pm	0.000 	$&$	\mathbf{0.366 	\pm	0.000} 	$&$	\mathbf{0.455 	\pm	0.000} 	$\\
			\hline\hline
			
			\textbf{Yale}	&	ACC		&		NMI		&		ARI		&		F1-score		&		Precision		&		Recall		&		Purity		\\ \hline	
			LT-MSC \cite{zhang2015low}	&$0.739\pm0.014$
			& $0.767\pm0.009$
			& $0.597\pm0.013$
			& $0.623\pm0.012$
			& $0.601\pm0.012$
			& $0.647\pm0.014$
			& $0.739\pm0.014$\\
			Proposed (LT-MSC)	&$	\mathbf{0.780 	\pm	0.015} 	$&$	\mathbf{0.784 	\pm	0.016} 	$&$	\mathbf{0.540 	\pm	0.037} 	$&$	\mathbf{0.572 	\pm	0.033} 	$&$	\mathbf{0.507 	\pm	0.049} 	$&$	\mathbf{0.659 	\pm	0.010} 	$&$	\mathbf{0.780 	\pm	0.015} 	$\\ \cdashline{1-8}
			ETLMC  \cite{TIPwu2019essential}	&$0.677\pm0.074$
			& $0.725\pm0.062$
			& $0.558\pm0.089$
			& $0.585\pm0.083$
			& $0.572\pm0.085$
			& $0.599\pm0.082$
			& $0.677\pm0.074$\\
			Proposed (ETLMC)	&$	\mathbf{0.733 	\pm	0.071} 	$&$	\mathbf{0.769 	\pm	0.058} 	$&$	\mathbf{0.617 	\pm	0.088} 	$&$	\mathbf{0.641 	\pm	0.082} 	$&$	\mathbf{0.628 	\pm	0.086} 	$&$	\mathbf{0.656 	\pm	0.077} 	$&$	\mathbf{0.733 	\pm	0.072} 	$\\
			\hline\hline

			\textbf{UCI}	&	ACC		&		NMI		&		ARI		&		F1-score		&		Precision		&		Recall		&		Purity		\\ \hline	
			LT-MSC \cite{zhang2015low}	&$	0.792 	\pm	0.009 	$&$	0.762 	\pm	0.009 	$&$	\mathbf{0.707 	\pm	0.014} 	$&$	\mathbf{0.737 	\pm	0.013} 	$&$	\mathbf{0.724 	\pm	0.012} 	$&$	0.749 	\pm	0.013 	$&$	0.809 	\pm	0.009 	$\\
			Proposed (LT-MSC)	&$	\mathbf{0.815 	\pm	0.003} 	$&$	\mathbf{0.765 	\pm	0.006} 	$&$	0.692 	\pm	0.007 	$&$	0.723 	\pm	0.007 	$&$	0.713 	\pm	0.008 	$&$	\mathbf{0.732 	\pm	0.005} 	$&$	\mathbf{0.815 	\pm	0.003} 	$\\ \cdashline{1-8}
			ETLMC  \cite{TIPwu2019essential}	&$	0.941 	\pm	0.023 	$&$	0.970 	\pm	0.013 	$&$	0.933 	\pm	0.029 	$&$	0.936 	\pm	0.027 	$&$	0.935 	\pm	0.0321 	$&$	0.938 	\pm	0.024 	$&$	0.942 	\pm	0.019 	$\\
			Proposed (ETLMC)	&$	\mathbf{0.997 	\pm	0.000} 	$&$	\mathbf{0.991 	\pm	0.000} 	$&$	\mathbf{0.992 	\pm	0.000} 	$&$	\mathbf{0.993 	\pm	0.000} 	$&$	\mathbf{0.993 	\pm	0.000} 	$&$	\mathbf{0.993 	\pm	0.000} 	$&$	\mathbf{0.997 	\pm	0.000} 	$\\
			\hline\hline

			\textbf{Coil20}	&	ACC		&		NMI		&		ARI		&		F1-score		&		Precision		&		Recall		&		Purity		\\ \hline	
			LT-MSC \cite{zhang2015low}	&$0.770\pm0.013$
			&$0.873\pm0.005$
			&$0.725\pm0.018$
			&$0.740\pm0.017$
			&${0.696\pm0.027}$
			&$0.790\pm0.005$
			&$0.817\pm0.005$\\
			Proposed (LT-MSC)	&$	\mathbf{0.844 	\pm	0.013} 	$&$	\mathbf{0.941 	\pm	0.003} 	$&$	\mathbf{0.828 	\pm	0.010} 	$&$	\mathbf{0.837 	\pm	0.010} 	$&$	\mathbf{0.774 	\pm	0.014} 	$&$	\mathbf{0.912 	\pm	0.008} 	$&$	\mathbf{0.880 	\pm	0.006} 	$\\
			 \cdashline{1-8}
			ETLMC  \cite{TIPwu2019essential}	&${0.956\pm0.037}	$
			&${0.977\pm0.012}	$
			&${0.950\pm0.035}	$
			&${0.952\pm0.033}	$
			&${0.937\pm0.049}	$
			&${0.969\pm0.019}	$
			&${0.965\pm0.027}$\\
			Proposed (ETLMC)	&$	\mathbf{0.988 	\pm	0.008} 	$&$	\mathbf{0.988 	\pm	0.004} 	$&$	\mathbf{0.977 	\pm	0.014} 	$&$\mathbf{	0.978 	\pm	0.013} 	$&$	\mathbf{0.975 	\pm	0.017} 	$&$	\mathbf{0.981 	\pm	0.009} 	$&$	\mathbf{0.988 	\pm	0.008} 	$\\
			\hline
			\hline
	\end{tabular}}
	\label{tab:otherGraph}
\end{table*}


  
%



\subsection{Visual Comparison of the Learned Similarity Matrix}

Fig. \ref{fig:visual} visualizes  the similarity matrices learned by two tensor based MVSC methods (i.e., LT-MSC and t-SVD-MSC), two TLRNs (i.e., SNN and TRPCA), and the proposed method. 
Compared with LT-MSC and t-SVD-MSC, the similarity matrix of our model shows a clear block-diagonal structure on ORL.  
On YaleB, there are many incorrect connections among samples from different clusters in the similarity matrices by  LT-MSC and t-SVD-MSC.  On the contrary, the connections of the similarity matrix of the proposed method are  sparsely distributed along the diagonal, indicating the majority of the connections are correct. 
The reason is that the proposed TLRN explicitly imposes a column-wise sparse regularization on the horizontal slices, such that it could remove the incorrect connections, while preserving the correct corrections by exploring the cross view information. 
Such observations also explain why the proposed method can achieve the excellent clustering performance over others in Tables \ref{tab:coil20}-\ref{tab:UCI-digital}.
In addition, compared with the two existing TLRNs, the recovered similarity matrices of SNN and TRPCA on both ORL and YaleB are quite dense.  Conversely, the similarity matrices of our method are sparse, which is very important for SC \cite{vonLuxburg2007}. The reason is that, unlike the existing TLRNs,  the proposed TLRN takes the unique characteristics of MVSC into account. 


\subsection{Parameter Analysis}
Fig. \ref{fig:para} shows
how the three hyper-parameters involved in our model affect clustering performance. On ORL, we can observe  that the highest ACC is achieved in a wide range of $\omega_1$ around $0.5$, i.e., $0.3\leq\omega_1\leq0.7$, 
indicating that the modelings of both  the  horizontal slices and the frontal slices make contributions  to the proposed model. The highest ACC of ORL  occurs when $\alpha\in[2,7]$, suggesting that the $\ell_{2,1}$ norm is more important than the low-rank term in the horizontal slices. The proposed model is also robust to $\lambda$ on ORL, i.e., the highest ACC occurs in a wide range of $\lambda$: $10\leq\lambda\leq 44$.
Moreover, on Coil20, the proposed model is also able to produce the highest ACC with a wide range of hyper-parameters. 
For the results shown in Tables \ref{tab:coil20}-\ref{tab:UCI-digital},  the selected hyper-parameters of the proposed model on all datasets  
are close to each other, 
and they all fall in the range where the highest ACC of ORL occurs,  i.e., $\omega_1\in[0.3,0.7],\alpha\in[2,7],\lambda\in[10, 44]$. All these observations validate that  the hyper-parameters of our method are relatively easy to choose, which proves the practicability of our model. 
\subsection{Ablation Study}
We also conducted ablation studies to comprehensively understand and evaluate the proposed model. First, we studied the effectiveness of the regularization terms contained in our model. As shown in Table \ref{tab:abS}, $\alpha=0$, $\omega_1=0$ and $\omega_2=0$ denote the proposed model without considering the column-wise sparsity, the frontal low-rankness, and the horizontal structure, respectively. From Table \ref{tab:abS}, we can see that the clustering performance   drops a lot when removing any regularization term contained in the proposed model, validating the effectiveness of each term. Specifically,  $\omega_1=0$ usually produces the worst clustering performance. The reason is that the frontal slice of the
tensor corresponds to the intra-view similarity relationship and  further takes responsible for the final clustering task. 

Second, as an initial similarity matrix is required in our method,  we investigated the adaptation  of the proposed model  to different similarity matrix construction methods. 
	Specifically, 
we used  the outputs of LT-MSC \cite{zhang2015low} and ETLMC \cite{TIPwu2019essential} as the  initial similarity matrices of our method, and the corresponding methods are denoted as Proposed (LT-MSC) and Proposed (ETLMC), respectively.  As shown in  Table \ref{tab:otherGraph}, it is obvious that Proposed (LT-MSC) (resp. Proposed (ETLMC)) can significantly improve the performance of LT-MSC (resp. ETLMC),  which validates the robustness of our model to different initial similarity matrix  construction methods. In other words,   the proposed model can act as  a post-processing technique to improve the performance of other existing  MVSC methods. 

\subsection{Comparison of Running Time}
Fig. \ref{fig:Running_time} illustrates the running time of all the methods on five datasets, where all of them were implemented with MATLAB on a Windows computer with a 3.7GHz Intel(R) i7-8700k CPU and 32.0 GB memory. We reported the average running time of 20 trails. 
From Fig. \ref{fig:Running_time}, we can observe that the running time of our method is roughly at the same level as  most methods under comparison, like SNN, TRPCA, and LT-MSC. Moreover, our method is much faster than MCLES on large datasets, including YaleB, Coil20 and UCI-digital. We can conclude that, compared with the state-of-the-art methods, our model is able to largely improve the clustering performance without increasing the running time. 


	\section{Conclusion and Future Work}
	In this paper, we have presented a novel tensor low-rank norm tailored  to MVSC, which explicitly  
	characterizes both the intra-view and inter-view of the multi-view samples. Based on that,  we formulated the MVSC as a low-rank tensor learning problem, and solved it with an augmented Lagrange multiplier method. 
	Our method is fundamentally  different from the existing methods which simply employ an existing TLRN that is designed for general purposes.
	The experimental results on five commonly-used benchmark datasets have substantiated the significant superiority of our model over eleven state-of-the-art MVSC methods, and validated the rationality of our modeling on  the special characteristics of MVSC  in low-rank tensor representation. 
	Moreover,  hyper-parameters of our model are relatively easy to determine, and our model is robust
	to different datasets and can adapt to different initial similarity matrices,  validating its potential in practice. 
	%
	We believe our perspective on MVSC will inspire this community.
	
	%
	In the future, we will 
	incorporate the proposed TLRN to a graph learning framework to learn a reasonable similarity matrix directly. 
	Second, the proposed LRTN has the potential to solve the ensemble clustering problem, where different data partitions of ensemble clustering can act as multiple views. 
	In addition, the final similarity matrix of our method used for clustering  is achieved by simply averaging the output tensor along the frontal direction, and thus it is expected that more elegant and effective fusion methods would further boost the performance. 

%
	\bibliographystyle{IEEEtran}
	\bibliography{acmbib,bib}
\end{document}